\documentclass[journal]{IEEEtai}

\usepackage[colorlinks,urlcolor=blue,linkcolor=blue,citecolor=blue]{hyperref}
\let\labelindent\relax
\usepackage{array}
\usepackage{graphicx}
\usepackage{amsmath,amssymb,amsfonts,amsthm,MnSymbol}
\usepackage{textcomp}
\usepackage{xspace}
\usepackage{cite}
\usepackage{picinpar}
\usepackage{url}
\usepackage{flushend}
\usepackage[latin1]{inputenc}
\usepackage{soul}
\usepackage{pifont}
\usepackage{color}
\usepackage{alltt,colortbl}
\usepackage{enumerate,enumitem}
\usepackage{siunitx}
\usepackage{breakurl}
\usepackage{epstopdf}
\usepackage{pbox}
\usepackage{booktabs,threeparttable,subfigure,multirow}
\usepackage{balance}
\usepackage{algorithmic}
\usepackage{textcomp}
\usepackage{ulem}
\usepackage{dcolumn}

\newcommand{\eg}{\textit{e}.\textit{g}.}
\newcommand{\T}{\mathrm{T}}
\newcommand{\D}{\mathrm{D}}
\newcommand{\RH}{\mathrm{H}}
\newcommand{\ie}{\textit{i}.\textit{e}.}

\newcommand{\tabincell}[2]{\begin{tabular}{@{}#1@{}}#2\end{tabular}}
\newcommand{\sig}{\rlap{$^*$}}
\newcommand*\mc[1]{\multicolumn{1}{c}{#1}}

\begin{document}
\title{\huge An Accurate and Interpretable Framework \\ for Trustworthy Process Monitoring}
\author{Hao Wang~\IEEEmembership{Student Member,~IEEE}, Zhiyu Wang, Yunlong Niu, Zhaoran Liu, Haozhe Li, Yilin Liao, Yuxin Huang, Xinggao Liu~\IEEEmembership{Member,~IEEE}
\thanks{Manuscript recieved Sept. 28, 2022. This work is supported by the National Key Research and Development Program of China (Grant No. 2021YFC2101100), the National Natural Science Foundation of China (11975207, 12075212, 12105246, 62073288),  and Zhejiang University NGICS Platform. } 
\thanks{Hao Wang, Zhiyu Wang, Yunlong Niu, Zhaoran Liu, Haozhe Li, Yilin Liao, Yuxin Huang, Xinggao Liu are with the State Key Laboratory of Industrial Control Technology, College of Control Science and Engineering, Zhejiang University, Hangzhou 310027, China (e-mail: haohaow@zju.edu.cn; onejoy@zju.edu.cn; nyl0405@zju.edu.cn;
22032057@zju.edu.cn; lihaozhe@zju.edu.cn; liaoyl@zju.edu.cn; hyx@zju.edu.cn; 
lxg@zju.edu.cn).
}
\thanks{Xinggao Liu is the corresponding author.}
\thanks{This paragraph will include the Associate Editor who handled your paper.}}

\markboth{Journal of IEEE Transactions on Artificial Intelligence, Vol. xx, No. x}
{Hao Wang \MakeLowercase{\textit{et al.}}: AttentionMixer}

\maketitle

\begin{abstract}
Trustworthy process monitoring seeks to build an accurate and interpretable monitoring framework, which is critical for ensuring the safety of energy conversion plant (ECP) that operates under extreme working conditions such as high pressure and temperature.
Contemporary self-attentive models, however, fall short in this domain for two main reasons. First, they rely on step-wise correlations that fail to involve physically meaningful semantics in ECP logs, resulting in suboptimal accuracy and interpretability. Second, attention matrices are frequently cluttered with spurious correlations that obscure physically meaningful ones, further impeding effective interpretation.
To overcome these issues, we propose AttentionMixer, a framework aimed at improving both accuracy and interpretability of existing methods and establish a trustworthy ECP monitoring framework.
Specifically, to tackle the first issue, we employ a spatial adaptive message passing block to capture variate-wise correlations. This block is coupled with a temporal adaptive message passing block through an \textit{mixing} operator,  yielding a multi-faceted representation of ECP logs accounting for both step-wise and variate-wise correlations.
Concurrently, to tackle the second issue, we employ a sparse message passing regularizer to filter out spurious correlations. 
We validate the efficacy of AttentionMixer using two real-world datasets from the radiation monitoring network for Chinese nuclear power plants.
\end{abstract}

\begin{IEEEImpStatement}

Given current operational challenges of ECPs, the imperative to institute a trustworthy monitoring framework is heightened. Contemporary monitoring solutions including Transformers manifest limitations in accuracy, interpretability, and overall trustworthiness. To fill this gap, our work introduces AttentionMixer for elevating the monitoring accuracy and interpretability. The proposed method is evaluated within a large-scale ECP monitoring project focusing on atmospheric nuclear radiation in China. The monitored dose rate data is made daily accessible via our project website\footnote{\url{https://data.rmtc.org.cn/gis/PubIndex.html}. We suggest accessing it using a virtual private network service through a server in mainland China.}.

\end{IEEEImpStatement}



\section{Introduction}
\label{sec:introduction}
\IEEEPARstart{M}{onitoring} energy conversion plants (ECPs) using high-dimensional input sensory data is a critical practice for ensuring energy safety and environmental security. 
With increasing demands for enhanced power conversion efficiency and carbon emission reduction, ECPs frequently operate under extreme conditions to approach the thermodynamic limit dictated by the Clausius theorem.
Examples of such extreme operating conditions include extremely high working fluid temperatures in coal-based power plants~\cite{wu2016embodied}, intense pressure from the force of falling water in hydroelectric plants, and high temperatures and pressures combined with nuclear radiation leak risks in nuclear power plants~\cite{safe1,safe2}. 
These extreme conditions exacerbate safety and environmental risks, underlining the need for a trustworthy monitoring approach to ensure energy safety and environmental security.

Trustworthiness in intelligent systems is increasingly crucial, with the growing concern of AI safety~\cite{liang2022advances,trustworthytai1,trustworthytai2}. 
Recent advancements mainly include interpretability, robustness, out-of-distribution generalization~\cite{wang2023out} and fairness~\cite{trustworthytai4,trustworthytai3,zhang2023robust,wang2022entire}, across domains like computer vision~\cite{nightingale2022ai}, natural language processing~\cite{sheth2021knowledge}, policy learning~\cite{DBLP:conf/icml/LiZCGLW23,DBLP:conf/kdd/LiZWKL023,wuite,wustable}, representation learning~\cite{DBLP:conf/kdd/YangCL0023,DBLP:conf/cvpr/YangLCSHW21}, causality~\cite{DBLP:conf/kdd/0001KCYGH023,DBLP:conf/kdd/00010YWXR0K22,DBLP:conf/sigir/WangZL0F023,wutest,DBLP:conf/iclr/LiLZ023} and recommendation~\cite{li2022multiple,li2023stabledr,wangescm,DBLP:conf/sigir/ZhangDCDT23,DBLP:conf/icml/LiXZ0023,DBLP:conf/sigir/YangWT23}.
We believe that such need for trustworthy solutions is especially pronounced for ECP monitoring systems, given the extreme conditions they operate under. In particular, a trustworthy ECP monitoring system should excel in both \textit{accuracy} and \textit{interpretability}~\cite{markus2021role,floridi2019establishing,liang2022advances}.
Here, \textit{interpretability} signifies the monitor's decision-making transparency, enabling experts to validate model's understanding of the data aligning with domain-specific knowledge. This feature serves as an essential diagnostic tool, reinforcing trust in the monitor's application in extreme settings.

Contemporary works focus on enhancing monitoring accuracy~\cite{yuan1,yuan2,xiao2022distributed}, fueled by advancements in measurement and learning techniques~\cite{li2022novel,fanlearnable,liu2023novel}. 
Initially, the landscape was dominated by traditional identification techniques such as ARIMA~\cite{Chandrakar2017}, alongside statistical methods like decision trees~\cite{xgb}, linear dynamic systems~\cite{li2023virtual,chenmode} and canonical variational models~\cite{chensvgd,dai2022incremental,dai2023variational}. 
The advent of deep learning has shifted the balance in favor of monitoring algorithms based on deep learning techniques~\cite{jiang2022data,huangbmoe}, exemplified by recurrent neural networks (RNN)~\cite{yuan1}, convolution neural networks (CNN)~\cite{wang2020remaining}, graph neural networks (GNN)~\cite{Wu2020}, each with distinct advantages and limitations. In particular, GNNs exploit variate-wise correlations for decision-making but necessitate a predefined adjacency matrix, often unavailable due to the scarcity of domain knowledge in ECP contexts. More recently, self-attentive models, a specialized subset of GNNs, have gained traction across diverse applications~\cite{jumper2021highly,chen2021knowledge}. These models obviate the need for predefined graph structures, making them particularly suited for intricate monitoring tasks like those in ECPs. Despite this potential, the application of self-attentive models in ECP monitoring remains unexplored, with conventional methods like CNNs and RNNs retaining their dominance in the realm of ECP monitoring.

This paper aims to explore the viability of self-attentive architectures for trustworthy ECP monitoring.
However, conventional self-attentive methods face shortcomings that hamper their trustworthiness as ECP monitors.
First, individual steps in time series such as ECP monitoring logs lack self-contained meaning\footnote{In contrast to natural language tokens, each of which has standalone semantics, individual data points in ECP logs are contextually reliant on temporal neighbors, often conveying information through patterns or trends.}, which renders step-wise correlations inadequate for discrimination and thus compromises accuracy.
Moreover, these step-wise correlations are difficult for domain experts to understand, negatively affecting interpretability.
Second, attention matrices tend to have high density, whereas the physically meaningful correlations (i.e., those based on the underlying physical principles of the ECP data-generation process, which can be readily understood and validated by domain experts) are typically sparse.
The spurious correlations overshadow the physically meaningful, expert-verifiable correlations, thereby compromising monitoring interpretability.

To handle these challenges, we propose AttentionMixer, a novel architecture built on a generalized message passing framework. To address the first issue, we incorporate a Spatial Adaptive Message Passing (SAMP) block, developed specifically to capture variate-wise correlations that are discriminative for ECP monitoring. These correlations, rooted in physical principles, not only make the model's decision process more comprehensible, but also allow it to be verifiable with expert knowledge. 
The SAMP block is combined with a Temporal Adaptive Message Passing (TAMP) block using a mixing operator, thereby creating a multi-faceted representation of ECP logs that accounts for both temporal and variate-wise correlations.
To tackle the second issue, we incorporate a sparse message passing regularizer (SMPR) designed to prune spurious correlations and accentuate the physically meaningful ones. 
By explicitly modeling variate-wise correlations and eliminating spurious correlations, it achieves better accuracy, interpretability, and overall trustworthiness in ECP monitoring.

\textbf{Contributions.} To summarize our main contributions:
\begin{itemize}
\item{We introduce AttentionMixer, a pioneering architecture that employs self-attentive mechanisms across both spatial and temporal dimensions, designed explicitly for achieving trustworthy ECP monitoring.}
\item{We develop a sparsification regularizer to prune spurious correlations, which enhances both the accuracy and interpretability of the ECP monitoring model. }
\item{We validate the accuracy and interpretability of AttentionMixer in real-world ECP monitoring practice.}
\end{itemize}

\textbf{Organization.}
In Section~\ref{sec:Background}, we introduce technical preliminaries requisite for understanding the proposed method.
In Section~\ref{sec:proposed}, we elaborate on the architectural components and learning objectives of AttentionMixer.
In Section~\ref{sec:experiment}, we deploy it in a practical ECP monitoring scenario, i.e., radiation dose rate monitoring around nuclear power plants, verifying its superior trustworthiness through empirical evaluation.
Conclusion and future work are discussed in Section~\ref{sec:conclusion}.


\section{Preliminaries}
\label{sec:Background}
\subsection{Message Passing Mechanism}\label{sec:mpnn}
Message passing is a general paradigm that encapsulates various spatial GNN models such as GCN, GraphSAGE, and GAT~\cite{ma2021deep}.
Let $\mathcal{M}$ and $\mathcal{U}$ be the message generation function and state update function, respectively; the k-th round of message passing for central node $\nu_i$ can be expressed as
\begin{equation}
\begin{aligned}
      m_i^k &= \sum_{\nu_j\in\mathbb{N_\mathcal{G}}(\nu_i)}{\mathcal{M}(F_i^k, F_j^k)},\\
      F_i^{k+1} &= \mathcal{U}(m_i^k, F_i^k),
\end{aligned}
\label{MPNN1}
\end{equation}
where $F_i^k$ is the feature embedding of $\nu_i$. In \eqref{MPNN1}, $\mathcal{M}$ generates the message embedding $m_i^k$ from the neighboring nodes of $\nu_i$ (denoted as $\mathbb{N}_\mathcal{G}(\nu_i)$), and $\mathcal{U}$ updates the feature embedding of $\nu_i$ by fusing the message embedding $m_i^k$ and the original feature embedding $F_i^k$. Despite the success of this paradigm in many fundamental fields, it is not directly applicable to ECP monitoring due to the lack of graph structure $\mathcal{G}$.

\subsection{Self-attentive Sequential Models}\label{sec:sam}
The self-attentive mechanism~\cite{Vaswani2017} allows node interactions through a synthesized attention matrix, without predefining graph structures, acting as a generalized message passing scheme. In time-series modeling, it facilitates the capture of long-term dependencies in emerging techniques.
For instance, FedFormer~\cite{zhou2022fedformer} employs a filter-enhanced attentive module to eliminate data noise; Informer~\cite{zhou2021informer} uses ProbSparse self-attention to remove redundant queries; Pyraformer~\cite{liu2021pyraformer} adopts pyramidal attention to capture multi-scale dependencies; LogTrans~\cite{li2019enhancing} combines convolution and self-attention to emphasize locality. Meanwhile, task-specific solutions have been developed for challenges such as multi-step forecasting~\cite{wu2020adversarial} and processing long history sequences~\cite{nie2022time}.

Despite the success of self-attentive architectures in fundamental fields, the prevalent reliance on step-wise correlations, which often contain noise and lack standalone physical meaning, impedes accuracy and interpretability in ECP monitoring.  Our study proposes a pioneering self-attentive architecture that captures both variate-wise and step-wise correlations, addressing this issue and highlighting the viability of self-attentive mechanisms in constructing trustworthy ECP monitors.


\section{Methodology}\label{sec:proposed}

In this section, we introduce AttentionMixer based on the message passing paradigm in Section~\ref{sec:mpnn}. 
The technical distinctions from the vanilla self-attentive method involve the explicit depiction of variate-wise correlations and the elimination of spurious relationships.

\subsection{Problem Definition}
We use bold uppercase letters (\eg, $\mathbf{X}$) to denote matrix, lowercase letters (\eg, $\mathbf{x}$) to denote associated vectors.
The input variables at the $t$-th step is denoted as $\mathbf{x}_t\in\mathbb{R}^{\D\times 1}$.
Given historical observations $\mathbf{X}_t:=\left[\boldsymbol{x}_{t+1},\boldsymbol{x}_{t+2},...,\boldsymbol{x}_{t+\T}\right]$ with window length $\T$, our goal is to learn a function $f:\mathbb{R}^\mathrm{T\times D}\rightarrow\mathbb{R}$ such that $f(\mathbf{X}_t)\rightarrow y_{t}^{\RH}$, where $y_{t}^{\RH}$ is the value of key performance indicators (KPIs) at the $t+\mathrm{T+H}$ step.

In the offline training stage, we aim to accurately forecast KPIs using historical KPIs and external factors~\cite{yuan1,yuan2}; in the inference stage, we anticipate that anomalous events will be indicated by large deviations between actual and forecast KPI values, as the training set only consists of normal logs.

\subsection{Spatial Adaptive Message Passing Module}\label{sec:samp}
\begin{figure*}
    \centering
    \subfigure[Overall architecture of AttentionMixer which mainly consists of the SAMP and TAMP blocks]{\includegraphics[width=0.96\textwidth, trim=0 850 350 0, clip]{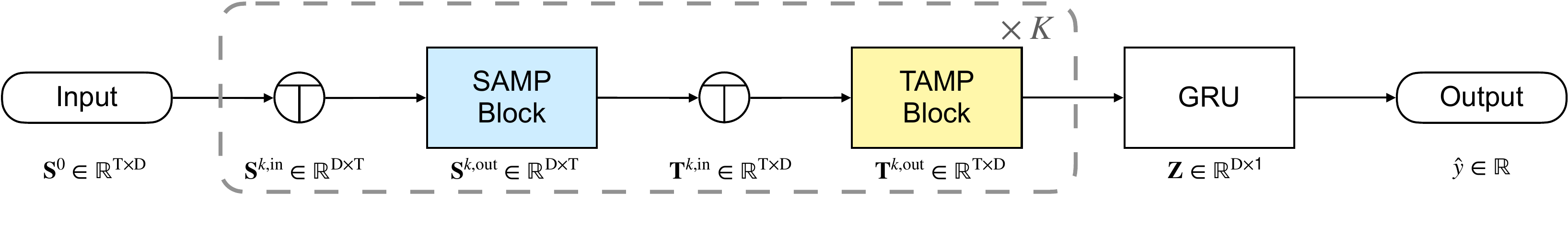}}
    \subfigure[Architecture of Spatial Adaptive Message Passing (SAMP) block]{\includegraphics[width=0.48\textwidth, trim=0 400 1150 200, clip]{images/DMPNv2.pdf}}
    \subfigure[Architecture of Temporal Adaptive Message Passing (TAMP) block]{\includegraphics[width=0.48\textwidth, trim=800 400 350 200, clip]{images/DMPNv2.pdf}}
    \caption{Overview of the architecture of AttentionMixer, where the output of each module is marked under the corresponding module.}
    \label{modelChart}
\end{figure*}

Canonical self-attention methods~\cite{Vaswani2017} treat each time step as a node and perform message passing at a temporal scale. This approach neglects physically meaningful variate-wise correlations and thus results in suboptimal accuracy and hinders interpretation in ECP monitoring. To address this issue, the SAMP module in this section depicts variate-wise correlations by performing message passing at a spatial scale, with each variate being treated as a node.
\autoref{modelChart} (b) shows the $k$-th round of message passing with input $\mathbf{S}^{k,\mathrm{in}}\in \mathbb{R}^{\D \times \T}$ and output $\mathbf{S}^{k,\mathrm{out}}\in \mathbb{R}^{\D \times \T}$.
The computation workflow is as follows.

\subsubsection{Message Generation} 
We first generate the outgoing message embeddings $\mathbf{V}_{\mathrm{S}}^{k} \in \mathbb{R}^{\D \times \T}$ for each node by feeding their state embeddings into a multilayer perception (MLP):
\begin{equation}
    \mathbf{V}_{\mathrm{S}}^{k} := \mathrm{MLP}(\mathbf{S}^{k,\mathrm{in}}),
\end{equation}
where $\mathbf{V}_{\mathrm{S}}^{k}\left[i\right]\in\mathbb{R}^\T$ is the outgoing message embedding of the i-th node, the MLP layer involves a learnable affine transformation and a linear activation function following~\cite{Vaswani2017}. 
We then define an attention matrix $\mathbf{M}_\mathrm{S}^k\in\mathbb{R}^{\D\times\D}$ as
\begin{equation}
    \begin{aligned}
        \mathbf{Q}_{\mathrm{S}}^{k} &:= \mathrm{MLP}(\mathbf{S}^{k,\mathrm{in}}), \quad
        \mathbf{K}_{\mathrm{S}}^{k} := \mathrm{MLP}(\mathbf{S}^{k,\mathrm{in}}),\\
        \mathbf{M}_\mathrm{S}^k &:= \mathrm{SoftMax}\left( \mathbf{Q}_{\mathrm{S}}^{k} * \left[\mathbf{K}_{\mathrm{S}}^{k}\right]^\T/\sqrt{\D} \right),
    \end{aligned}
\end{equation}
where $\mathbf{M}_\mathrm{S}^k\left[ i,j \right]$ is a generalized similarity measure between the i-th node and the j-th node;
the division of $\sqrt{\D}$ avoids numerical errors during optimization~\cite{Vaswani2017}.
The SoftMax operation normalizes the attention scores of each node~\cite{Vaswani2017}.

Finally, we generate the incoming message embeddings $\mathbf{U}_{\mathrm{S}}^{k} \in \mathbb{R}^{\D \times \T}$ for each node by aggregating the outgoing message embeddings using the attention matrix $\mathbf{M}_\mathrm{S}^k$.
The main intuition here is straightforward: neighbors with more remarkable similarities should convey more important messages.
As such, the incoming message embeddings are calculated as
\begin{equation}
    \mathbf{U}_\mathrm{S}^k := \mathbf{M}_\mathrm{S}^k * \mathbf{V}_\mathrm{S}^k,
\end{equation}
where the incoming message embedding of the i-th node is a weighted sum of the outgoing message embeddings of its neighbors, with the weight being its attention to its neighbors:
\begin{equation}
    \mathbf{U}_\mathrm{S}^k \left[ i,: \right]:= \sum_{j=1}^{\D} \mathbf{M}_\mathrm{S}^k\left[ i,j \right]\mathbf{V}_\mathrm{S}^k\left[ j,:\right].
\end{equation}

\subsubsection{State Update} The state embedding for each node is renewed according to its previous state embedding $\mathbf{S}^{k,\mathrm{in}}$ and incoming message embedding $\mathbf{U}_\mathrm{S}^k$ as
\begin{equation}
    \mathbf{S}^{k,\mathrm{out}} := \mathrm{LayerNorm}\left( \mathbf{S}^{k,\mathrm{in}}+\mathbf{U}_\mathrm{S}^k \right),
\end{equation}
where the residual link preserves the previous state embeddings that are excluded in the generation process of $\mathbf{U}_\mathrm{S}^k$; the LayerNorm technique standardizes the data distribution at each layer to stabilize the training process.

\subsection{Temporal Adaptive Message Passing Module}\label{sec:tamp}
The SAMP module focuses on variate-wise correlations while ignoring temporal correlations. However, temporal correlations are crucial for capturing auto-regressive characteristics in monitoring logs, necessitating the TAMP module.

\autoref{modelChart} (c) illustrates the $k$-th round of temporal message passing, with $\mathbf{T}^{k,\mathrm{in}}\in \mathbb{R}^{\T \times \D}$ representing the input and $\mathbf{T}^{k,\mathrm{out}}\in \mathbb{R}^{\T \times \D}$ representing the output. The computation workflow is similar to that of SAMP, but TAMP relies on correlations between time steps rather than variates. TAMP models such step-wise correlations by performing message passing at a temporal scale, with each time step being treated as a node and the input $\mathbf{T}^{k,\mathrm{in}}$ serving as the state embeddings of $\T$ nodes. The message passing route $\mathbf{M}_\T^k\in\mathbb{R}^{\T\times \T}$ thus depends on the correlations between time steps.

\subsection{Sparse Message Passing Regularizer}\label{sec:smpr}
The correlations learned by self-attentive models tend to have high density according to \autoref{explainChart} (a) , whereas the actual relationships in the physical world are sparse, as per the Occam's razor principle. This implies the existence of many spurious relationships, which can lead to overfitting and obscure physically meaningful relationships~\cite{bigbird}, ultimately compromising the accuracy and interpretability of monitoring. To address this issue, several graph sparsification techniques~\cite{chen2021knowledge, zhou2021informer} have been proposed to obtain interpretable and meaningful graph structures. However, most of these techniques rely on sampling, resulting in significant variance in both training and inference phases, which is undesirable for ECP monitoring under extreme working conditions.

To avoid introducing sampling during model inference, regularization of the graph structure during the training process is preferable. In our framework, the graph structure is represented by the message passing route $\mathbf{M}$, and a natural approach is to regularize its $\ell$-1 norm as follows:
\begin{equation}
\begin{aligned}
    \mathcal{L}_{\mathrm{smpr}}&:=\sum_{k=1}^\mathrm{K}\left\|\mathbf{M}_\mathrm{S}^k\right\|_{\mathrm{1}} + \sum_{k=1}^\mathrm{K}\left\|\mathbf{M}_\mathrm{T}^k\right\|_{\mathrm{1}}, \\   \left\|\mathbf{M}_\mathrm{S}\right\|_{\mathrm{1}}&:=\sum_{i,j=1}^{\mathrm{D}}\left|\mathbf{M}_\mathrm{S}\left[i,j\right]\right|, \quad\left\|\mathbf{M}_\mathrm{T}\right\|_{\mathrm{1}}:=\sum_{i,j=1}^{\mathrm{T}}|\mathbf{M}_\mathrm{T}[i,j]|
\end{aligned}   
\end{equation}
where $\Vert \cdot \Vert_1$ indicates the mean absolute value of the matrix, which reinforces sparse weights during training. This regularization helps eliminate spurious relationships and highlight physically meaningful relationships, ultimately improving the accuracy and interpretability of monitoring.

\subsection{AttentionMixer Framework for ECP Monitoring}\label{sec:dampn}
\subsubsection{Overall Architecture}
As depicted in \autoref{modelChart} (a), AttentionMixer consists of a cascade of SAMP and TAMP blocks. The preprocessed data is first fed into SAMP, \ie, $\mathbf{S}^{1,\mathrm{in}}=\mathbf{S}^0$, where messages are passed through the adaptive spatial graph.
Next, a mixer operator transposes the spatial and temporal dimensions of SAMP's output to get TAMP's input:
\begin{equation}
    \begin{aligned}
        \mathbf{S}^{k,\mathrm{out}}&:=\mathrm{SAMP}\left( \mathbf{S}^{k,\mathrm{in}} \right), \\
        \mathbf{T}^{k,\mathrm{in}}&:=\mathrm{Transpose}\left( \mathbf{S}^{k,\mathrm{out}} \right),\\
    \end{aligned}
\end{equation}
where $k\leq\mathrm{K}$ is the round of current message passing.
Afterwards, TAMP extracts the temporal correlations in the data, and the transpose of its output is the input to the SAMP block in the next round of message passing:
\begin{equation}
    \begin{aligned}
        \mathbf{T}^{k,\mathrm{out}}&:=\mathrm{TAMP}\left( \mathbf{T}^{k,\mathrm{in}} \right),\\
        \mathbf{S}^{k+1,\mathrm{in}}&:=\mathrm{Transpose}\left( \mathbf{T}^{k,\mathrm{out}} \right).
    \end{aligned}
\end{equation}
The output of the last TAMP block, $\mathbf{T}^{K,\mathrm{out}}$, is fed into a gated recurrent unit (GRU) decoder to get the dose rate estimate:
\begin{equation}
\label{yPred}
    \mathbf{Z}:=\mathrm{GRU}\left( \mathbf{T}^{\mathrm{K},\mathrm{out}} \right), \quad
    \hat{y}_{t}^{\RH}:=\mathrm{Linear}\left(\mathbf{Z}  \right),
\end{equation}
where $\hat{y}_{t}^{\RH}$ is the output of AttentionMixer, $\RH$ is the forecast horizon, $\mathrm{Linear}(\cdot)$ is an MLP-layer with linear activation.
\subsubsection{Learning Objective}
The learning objective aims to minimize the forecast error and the density of message passing.
Given input $\mathbf{X}_t$ and label $y_{t}^{\RH}$, the forecast error is
\begin{equation}
\label{sparseReg}
    \mathcal{L}_{\mathrm{pred}}:=\left(y_{t}^{\RH}-\hat{y}_{t}^{\RH}\right)^2,
\end{equation}
where $y_{t}^{\RH}$ is the actual radiation dose rate at the $t+\RH$ time step,
$\hat{y}_{t}^{\RH}$ is the output of the proposed AttentionMixer given input $\mathbf{S}^0:=\mathbf{X}_t$.
The overall learning objective is
\begin{equation}
\label{loss}
    \mathcal{L}_\mathrm{Mixer}=\mathcal{L}_{\mathrm{pred}}+\lambda\cdot\mathcal{L}_\mathrm{smpr},
\end{equation}
where $\lambda$ controls the strength of SMPR.
The network structure and learning objective cooperate to improve accuracy and interpretability, both critical for nuclear radiation monitoring.

\section{Experiments}\label{sec:experiment}
To demonstrate the efficacy of AttentionMixer for ECP monitoring, four aspects deserve empirical investigation. The purpose of this section is to present these aspects individually:
\begin{enumerate}[leftmargin=*]
    \item \textbf{Performance:} \textit{Does AttentionMixer work?} Section \ref{sec:overall} compares the overall performance of AttentionMixer against various baselines in the realms of time series forecast and process monitoring, using two collected real-world logs for nuclear ECP monitoring.
    \item \textbf{Gains:} \textit{Why does it work?} Section \ref{sec:discuss} deconstructs it to identify the sources of performance gain relative to the vanilla self-attention mechanism. The interpretability is also explored through visualizing the attention scores.
    \item \textbf{Sensitivity:} \textit{Is AttentionMixer sensitive to hyperparameters?} Section \ref{sec:param} provides insights into the selection of hyperparameters and demonstrates its superiority over various baselines despite variations in some hyperparameters.
    \item \textbf{Complexity:} \textit{Is its complexity feasible in practice?} Section \ref{sec:complex} analyzes the complexity of SAMP and TAMP modules and reports actual running time in various settings.
\end{enumerate}

\subsection{Background on Nuclear Radiation Monitoring}
\begin{figure}
    \centering
    \subfigure[]{\includegraphics[scale=0.6, trim=20 0 2 5,clip]{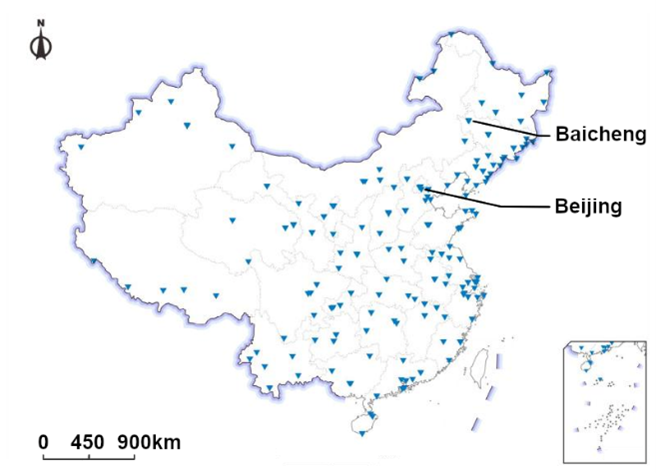}}
    \subfigure[]{\includegraphics[scale=0.4, trim=20 10 52 43,clip]{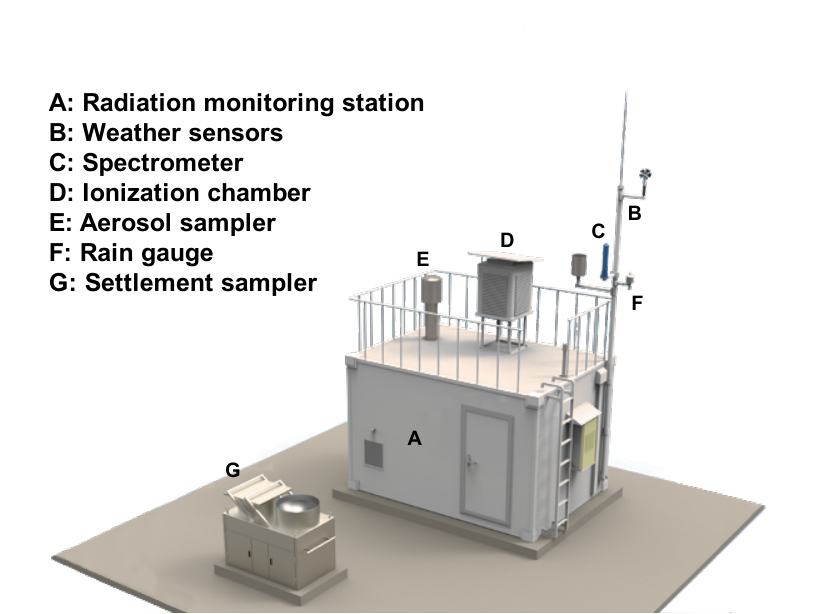}}
    \caption{Overview of the national automatic nuclear radiation monitoring network in China: (a) the distribution of monitoring stations that have been deployed; (b) the composition of a specific monitoring station.}
    \label{stationChart}
\end{figure}
\begin{table}
    \setlength{\tabcolsep}{3pt}
    \caption{Variable Definitions for the Monitoring Log}
    \label{dataTab}
    
      \begin{tabular}{p{130pt} p{110pt}}
        \toprule
        \textbf{Field}               & \textbf{Unit of measure}\\
        \midrule
        \textbf{Atmospheric radiation}\\
        Dose rate           & nGy / h\\
        Spectrum (1024 channels)       & ANSI/IEEE N42.42\\
        \midrule
        \textbf{Meteorological conditions}\\
        Temperature           & $^{\circ}$C\\
        Humidity       & \%\\
        Atmosphere pressure           & hPa\\
        Wind direction       & Clockwise angle\\
        Wind speed           & m / s\\
        Precipitation indicator       & Boolean value\\
        Amount of precipitation      & mm\\
        \midrule
        \textbf{Spectrometer operating conditions}\\
        Battery voltage     & V\\
        Spectrometer voltage        & V\\
        Spectrometer temperature         & $^{\circ}$C\\
        \bottomrule
      \end{tabular}
    
\end{table}

As a resilient and reliable source of energy~\cite{use1,use2}, nuclear power plants generate nearly 800 billion kilowatt-hours of electricity per year in America, supplying more than 60\% of the emission-free electricity, reducing about 500 million metric tons of carbon emissions~\cite{roth2017going}.
However, as an extreme-condition ECP, it raises considerable concerns regarding environmental and energy security~\cite{safe1,safe2}. In nuclear anomalies, excessive radionuclides are released into the atmosphere, causing severe environmental pollution.
As a preventive technology, automatic monitoring networks for nuclear radiation have been widely deployed, such as
the Atmospheric Nuclear Radiation Monitoring Network (ANRMN) in China, comprising around 500 individual nuclear radiation monitoring stations, as shown in \autoref{stationChart} (a).
The key performance indicator (KPI) to monitor is the $\gamma$-ray dose rate in the atmosphere~\cite{weather1}, which indicates radionuclide intensity.
However, dose rate is influenced not only by radionuclide intensity but also by meteorological and spectrometer operating conditions~\cite{weather1,weather3}.
For example, radionuclide radon in the air can be washed to the ground during rainfall, leading to a pseudo increase in the recorded dose rate~\cite{weather1}.
Ignoring such external factors can lead to false alarms and unreliable monitoring.

\begin{figure}
\centering
\includegraphics[width=0.442\columnwidth]{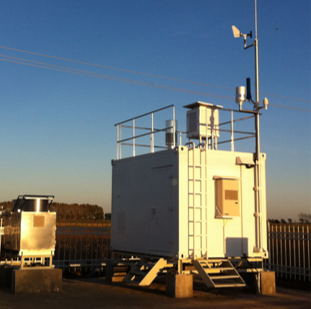}
\includegraphics[width=0.54\columnwidth]{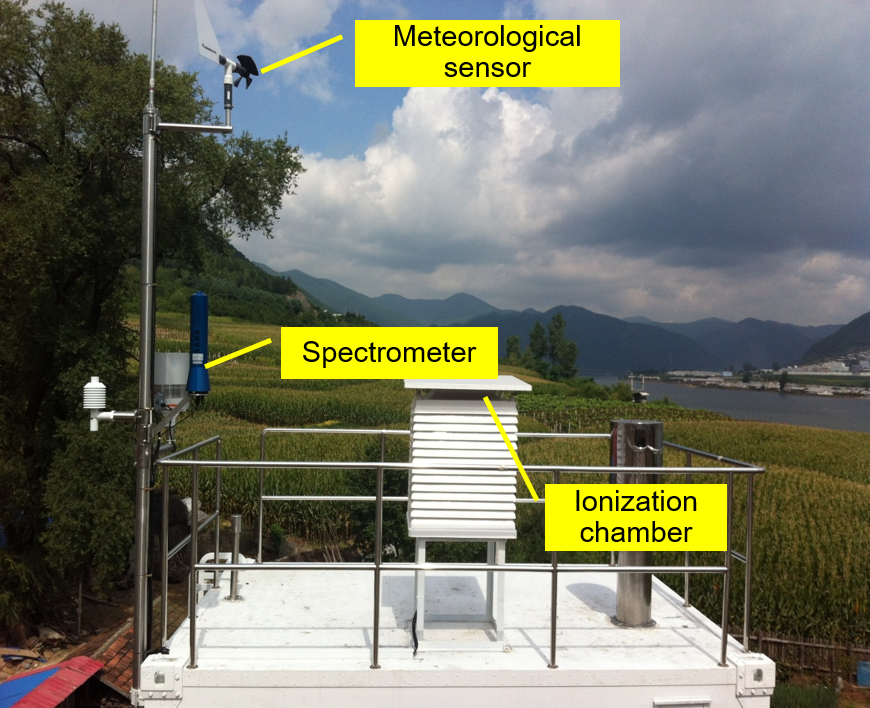}
\caption{Real-world monitoring stations with key sensors being highlighted.}\label{fig:real}
\end{figure}

\begin{figure}
    \centering
    \subfigure[]{\includegraphics[width=4.7cm, trim=5 5 245 90,clip]{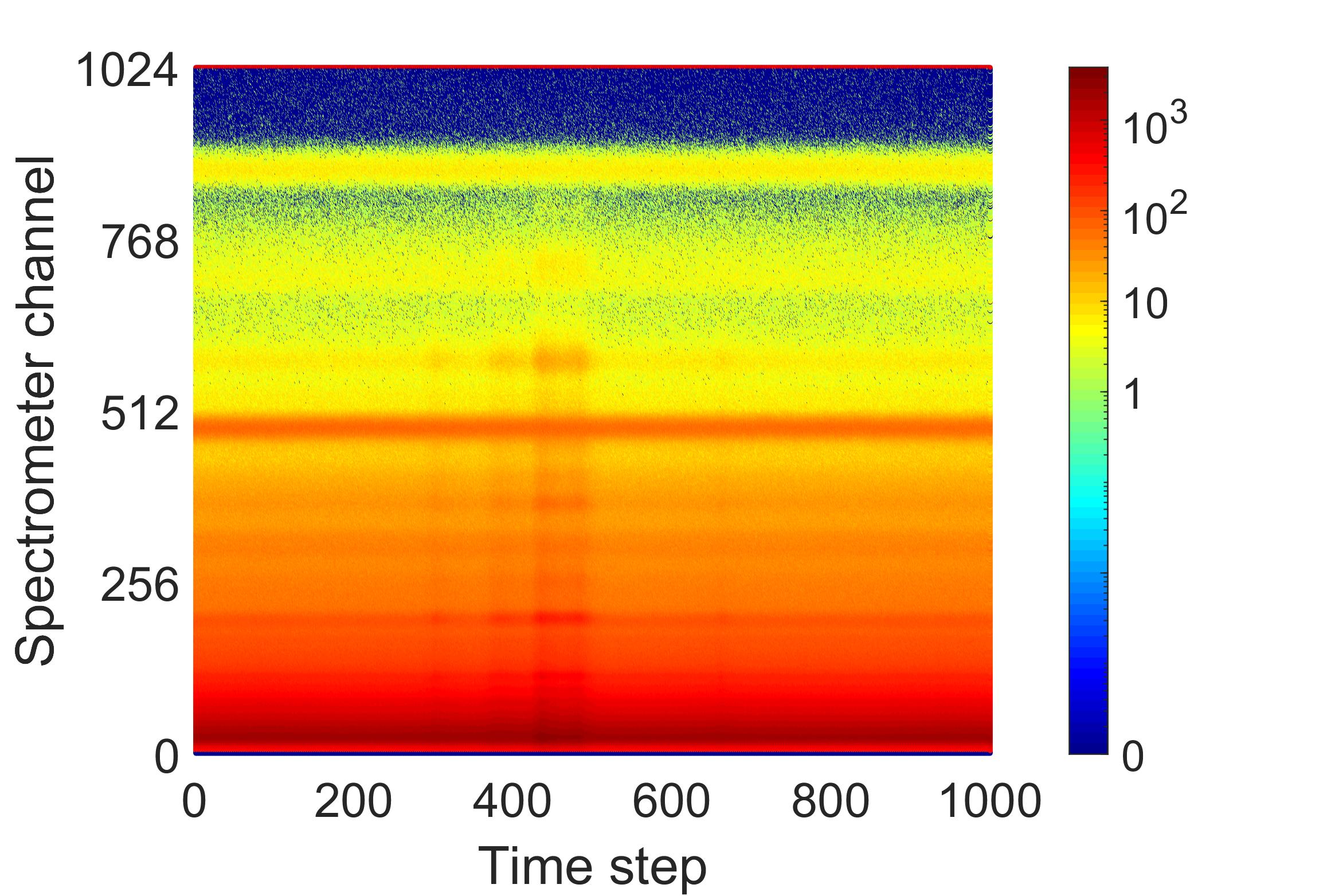}}
    \subfigure[]{\includegraphics[width=4cm, trim=215 350 218 340,clip]{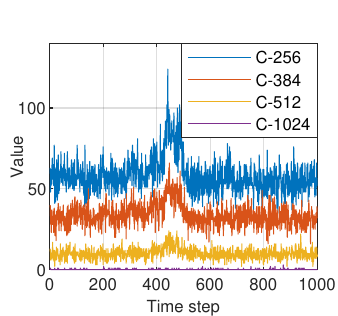}}
    \caption{Spectrometer measurements and dose rates of 1000 sequential time steps starting from 2020/5/15 21:54:00, in Baicheng Station, Jilin.}
    \label{dataFig}
\end{figure}

To construct a data-driven ECP monitoring system, we mainly consider historic $\gamma$-ray dose rate measured by ionization chamber in \autoref{fig:real}.
Another important factor is the spectrum measured by spectrometers\footnote{The spectrometer employs the SARA detector by ENVINET GmbH.}, offering detailed insights into environmental radionuclide concentrations.
\autoref{dataFig} (a) shows a 1024-channel spectrum with different channels exhibiting varying patterns and the energy decaying in higher channels.
\autoref{dataFig} (b) provides further details on the dynamics of the measured spectrum with four selected channels, which are non-stationary due to the influence of external factors. To calibrate such external impacts, meteorological conditions (\eg, precipitation) and spectrometer operating conditions (\eg, battery voltage) are supplemented in the monitoring logs.
    
\subsection{Experimental Setup}
\begin{table}
    \centering
\caption{Description of sampling strategy.}\label{dataTab2}
\resizebox{\columnwidth}{!}{
\begin{tabular}{llllll}
\toprule
Station & Location & \#Sample & \#Variable & Interval & Start time \\\midrule
Changchi & Beijing & 38,686 & 1,034 & 5 min & 10/3/2020\\
Industrial Park& Baicheng & 38,687 & 1,034 & 5 min & 21/4/2020\\
\bottomrule
\end{tabular}
  }
\end{table}
\subsubsection{Datasets} Two industrial datasets are constructed using the monitoring logs obtained from the ANRMN project. These datasets consist of 1034 input variables, as described in \autoref{dataTab}, which includes the 1024-channel spectrum, 7-channel meteorological conditions, and 3-channel spectrometer operating conditions. The sampling strategy used to create both datasets is detailed in \autoref{dataTab2}.
Both datasets are divided chronologically into training, validation, and test sets in a ratio of 0.7:0.15:0.15. Finally, they are scaled using a min-max scaler to facilitate model training and evaluation.

\subsubsection{Baselines} AttentionMixer is compared with three groups of baselines.
(1) Identification-based models: Auto-Regression (AR)~\cite{Chandrakar2017}, Moving Average (MA)~\cite{Chandrakar2017}, AutoRegressive Integrated Moving Average (ARIMA)~\cite{Chandrakar2017};
(2) Statistic-based models:  Lasso Regression (LASSO)~\cite{Chan2017}, Support Vector Regression (SVR)~\cite{Tang2012}, Random Forest (RF)~\cite{rf}, Gradient Boosting Decision Tree (GBDT)~\cite{xgb}, eXtreme Gradient Boosting (XGB)~\cite{xgb};
(3) Deep models: Long Short-Term Memory (LSTM)~\cite{Choi2020}, Gated Recurrent Unit (GRU)~\cite{Zhang2018}, Transformer~\cite{Vaswani2017}, Informer~\cite{zhou2021informer}, FedFormer~\cite{zhou2022fedformer}. Consistent with AttentionMixer, we utilize GRU as the decoder for former-like baselines for fair comparison.

\subsubsection{Training Strategy} Deep models are trained for 200 epochs with early stopping strategy. We set the batch size to 64, the window length $\mathrm{T}$ to 16, the learning rate to 0.001, and the round $\mathrm{K}$ of message passing to 2. The strength $\lambda$ of SMPR is set to $5e^{-5}$ for the Baicheng station and $1e^{-5}$ for the Beijing station, respectively. 
We adhere this setup for fair comparison. 
We check-point models on the validation set every epoch, conduct early stopping strategy with patience 15, and report the performance upon hitting the stopping criteria.

\subsubsection{Evaluation Strategy}
The coefficient of determination ($\mathrm{R}^2$) is used as the main fitness metric in line with~\cite{Wu2020}.
\begin{equation}
    \mathrm{R}^{2} =1-\sum_{t=1}^{\mathrm{N}}\left(y_{t}^\RH-\hat{y}_{t}^\RH\right)^{2} / \sum_{i=1}^{\mathrm{N}}\left(y_{t}^\RH-\bar{y}_{t}^\RH\right)^{2},
\end{equation}
where $y_{t}^\RH$ is the actual dose rate, $\hat{y}_{t}^\RH$ is the estimated dose rate, $\bar{y}_{t}^\RH$ is the mean value of $y_{t}^\RH$ for $t=1:\mathrm{N}$, $\mathrm{N}$ is the size of test set.
To provide a comprehensive model comparison, root mean squared error (RMSE) and mean absolute error (MAE) are employed as additional metrics.

\subsection{Overall Performance}\label{sec:overall}
\begin{figure}
\centering
\subfigure[Error on the validation set.]{
\includegraphics[width=0.48\columnwidth]{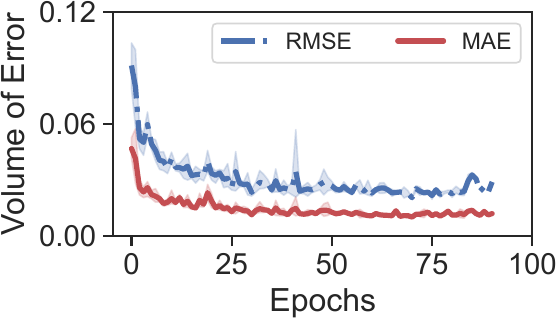}}
\subfigure[Accuracy on the eval set.]{
\includegraphics[width=0.48\columnwidth]{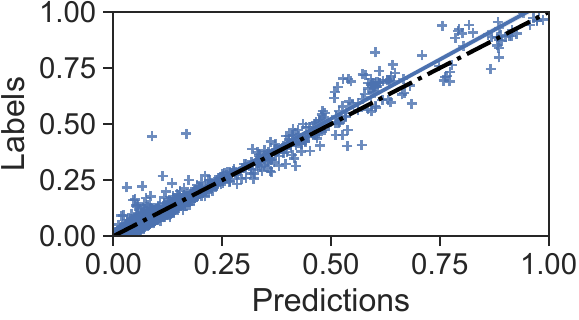}}
\caption[Caption for LOF]{Monitoring performance of AttentionMixer on Beijing dataset.}\label{fig:result}
\end{figure}

\begin{table*}
\setlength{\tabcolsep}{9pt}
\centering
\caption{Comparative Study on the Beijing and Baicheng datasets over four forecast horizons.}
\label{comparativeTab}
\begin{threeparttable}
\begin{tabular}{lllccccccccc}
\toprule
\multicolumn{2}{c}{\multirow{2}{*}{Methods}} & \multirow{2}{*}{Metrics} & \multicolumn{4}{c}{\tabincell{l}{Beijing Station}} && \multicolumn{4}{c}{\tabincell{l}{Baicheng Station}} \\
\cmidrule{4-7}\cmidrule{9-12}
& & & \mc{H=1} & \mc{H=3} & \mc{H=5} & \mc{H=7} && \mc{H=1} & \mc{H=3} & \mc{H=5} & \mc{H=7}\\ \midrule
\multirow{6}{*}{\tabincell{l}{Identification \\  based models}} &\multirow{2}{*}{AR}        &MAE    &0.0925 &0.1005 &0.1068 &0.1161            &&0.0999 &0.1329 &0.1669 &0.1899\\
                                                                    &                       &$\mathrm{R}^2$  &0.7672 &0.7265 &0.6887 &0.6455            &&0.7051 &0.5110 &0.2196 &-0.0282\\
                                                                    &\multirow{2}{*}{MA}    &MAE    &0.0952 &0.1027 &0.1081 &0.1161            &&0.1537 &0.1794 &0.2019 &0.2201\\
                                                                    &                       &$\mathrm{R}^2$  &0.7563 &0.7193 &0.6828 &0.6437            &&0.3457 &0.1095 &-0.1363&-0.3642\\
                                                                    &\multirow{2}{*}{ARIMA} &MAE    &0.0962 &0.0982 &0.1048 &0.1000            &&0.0818 &0.1094 &0.1452 &0.1762\\
                                                                    &                       &$\mathrm{R}^2$  &0.7700 &0.7339 &0.7228 &0.7342            &&0.7954 &0.6611 &0.3940 &0.1258\\ \midrule
\multirow{10}{*}{\tabincell{l}{Statistic \\ based models}}             &\multirow{2}{*}{LASSO} &MAE    &0.0500 &0.0511 &0.0527 &0.0546            &&0.0207 &0.0210 &0.0216 &0.0224\\
                                                                    &                       &$\mathrm{R}^2$  &0.4015 &0.3663 &0.3124 &0.2491            &&0.3998 &0.3708 &0.3260 &0.2719\\
                                                                    &\multirow{2}{*}{SVR}   &MAE    &0.0723 &0.0712 &0.0677 &0.0664            &&0.0486 &0.0599 &0.0647 &0.0655\\
                                                                    &                       &$\mathrm{R}^2$  &0.4743 &0.4843 &0.4828 &0.4429            &&0.3247 &-0.0073 &-0.2074 &-0.2936\\
                                                                    &\multirow{2}{*}{RF}   &MAE     &0.0337 &0.0360 &0.0383 &0.0413            &&0.0159 &0.0171 &0.0172 &1.0183\\
                                                                    &                       &$\mathrm{R}^2$  &0.8580 &0.8193 &0.7589 &0.6753            &&0.8918 &0.8393 &0.7879 &0.7118\\
                                                                    &\multirow{2}{*}{GBDT}   &MAE    &0.0396 &0.0406 &0.0420 &0.0444            &&0.0176 &0.0175 &0.0178 &0.0183\\
                                                                    &                       &$\mathrm{R}^2$  &0.7453 &0.7137 &0.6543 &0.5678            &&0.7840 &0.7559 &0.7057 &0.6460\\
                                                                    &\multirow{2}{*}{XGB}   &MAE    &0.0302 &0.0332 &0.0370 &0.0412            &&0.0105 &0.0118 &0.0136 &0.0149\\
                                                                    &                       &$\mathrm{R}^2$  &0.8943 &0.8393 &0.7684 &0.6719            &&0.9485 &0.9012 &0.8335 &0.7555\\ \midrule
\multirow{8}{*}{\tabincell{l}{Deep models}}      &\multirow{2}{*}{LSTM}  &MAE & 0.0181 & 0.0214 & 0.0251 & 0.0294 &  & 0.0160 & 0.0170 & 0.0183 & 0.2000 \\
                                                                    &                       &$\mathrm{R}^2$  & 0.9025 & 0.8343 & 0.7478 & 0.6547 &&0.8916 & 0.8331 & 0.7640 & 0.6886\\
                                                                    &\multirow{2}{*}{GRU}   &MAE    &0.0177 & 0.0214 & 0.0247 & 0.0280 && 0.0179 & 0.0189 & 0.0198 & 0.0210 \\
                                                                    &                       &$\mathrm{R}^2$  & 0.9021 & 0.8382 & 0.7577 & 0.6742 && 0.8621 & 0.8046 & 0.7426 & 0.6720\\ 
                                                                    &\multirow{2}{*}{Transformer}   &MAE    & 0.0142 & 0.0223 & 0.0282 & 0.0334 && 0.0111 & 0.0154 & 0.0177 & 0.0210 \\
                                                                    &                       &$\mathrm{R}^2$  & 0.9642 & 0.8640 & 0.7915 & 0.6826 && 0.9423 & 0.7996 & 0.7318 & 0.6252\\ 
                                                                    &\multirow{2}{*}{Informer}   &MAE    & 0.0189 & 0.0287 & 0.0276 & 0.0482 &&0.0123 & 0.0140 & 0.0221 & 0.0186 \\
                                                                    &                       &$\mathrm{R}^2$  & 0.9082 & 0.8444 & 0.7745 & 0.6144 && 0.8949 & 0.8416 & 0.6941 & 0.6738\\ 
                                                                    &\multirow{2}{*}{Fedformer}   &MAE    & \uline{0.0104} & \textbf{0.0185} & \uline{0.0233} & \textbf{0.0254} && \uline{0.0069} & \uline{0.0100} & \uline{0.0125} & \textbf{0.0149}\\
                                                                    &                       &$\mathrm{R}^2$  & \uline{0.9750} & \textbf{0.9239} & \textbf{0.8506} & \textbf{0.7684} && \uline{0.9765} & \uline{0.9215} & \uline{0.8483} & \textbf{0.7882}\\ 
                                                                    \midrule
\multirow{2}{*}{\tabincell{l}{Current work}}                        &\multirow{2}{*}{AttentionMixer} &MAE    & \textbf{0.0104} & \uline{0.0197} & \textbf{0.0213} & \uline{0.0280} && \textbf{0.0066} & \textbf{0.0093} & \textbf{0.0112} & \uline{0.0156}\\
                                                                    &                       &$\mathrm{R}^2$  &\textbf{0.9756} & \uline{0.9044} & \uline{0.8465} & \uline{0.7245} && \textbf{0.9783} & \textbf{0.9229} & \textbf{0.8708} & \uline{0.7859}\\
\bottomrule

\end{tabular}
    \begin{tablenotes}
        \footnotesize
        \item Bold fonts and underlined fonts indicate the best and second best performance for each metric, respectively.
    \end{tablenotes}
\end{threeparttable}
\end{table*}

The comparative analysis provided in \autoref{comparativeTab} explores the performance of AttentionMixer and other competing models at four distinct forecast horizons ($\RH=1,3,5,7$). The observations can be summarized as follows:

Identification-based models like ARIMA exhibit practical short-term forecasting capabilities, achieving an $\mathrm{R}^2$ of 0.77 and 0.79 on the Beijing and Baicheng datasets, respectively. 
It indicates the existence of autocorrelation in the data. However, these models are primarily designed to handle linear autocorrelation, leaving them ill-equipped to handle non-linear autocorrelation prevalent in many datasets. 
This limitation becomes evident in the long-term forecast performance, with ARIMA achieving an $\mathrm{R}^2$ of only 0.12 on the Baicheng dataset for H=7, primarily due to strong non-linear autocorrelation.

Statistic-based models incorporate external factors such as meteorological conditions, achieving competitive short-term forecasting performance. Notably, non-linear estimators like RF, GBDT, and XGB, with their higher modeling capacity, outperform their linear models (LASSO, SVR). For instance, RF surpasses LASSO and SVR with a MAE of 0.033 and 0.0159. Moreover, XGB's superior performance, even over traditional deep models like LSTM and GRU on both datasets, underscores the importance of effectively capturing variate-wise intersections in improving ECP monitoring accuracy.

Deep models achieve the best overall performance among all baselines. RNN-based deep models, such as LSTM and GRU, outperform all non-deep models except XGB. Self-attentive models further enhance monitoring performance significantly, indicating their potential for ECP monitoring. Interestingly, the Informer model falls short in comparison to Transformer, possibly due to the overly aggressive sparsification strategy it employs, which potentially diminishes model performance. A similar trend is observed in AttentionMixer, where excessive SMPR strength can reduce performance.

AttentionMixer performs outstandingly compared with most baseline models across various metrics and datasets. Its superior accuracy is attributed to its explicit modeling of variate-wise correlations and effective elimination of spurious correlations. Notably, while FedFormer sometimes achieves competitive or even superior performance, it bypasses the explicit modeling of variate-wise correlations. As a result, its decision-making process remains black-box and unverifiable by domain experts, thereby reducing its transparency and trustworthiness for ECP monitoring.

Finally, \autoref{fig:result} offers a more in-depth exploration of AttentionMixer's performance. The model converges swiftly during the training phase, with the validation error stabilizing after around 25 epochs. In the testing phase, AttentionMixer accurately estimates the dose rates, as indicated by the close alignment of estimated and actual values.

\subsection{Discussion on Trustworthiness}\label{sec:discuss}
As discussed in Section \ref{sec:introduction}, trustworthy ECP monitoring framework should be both accurate and interpretable. In this section, we conduct empirical studies to investigate these aspects for AttentionMixer.
\subsubsection{Accuracy}
\begin{table*}[]
\setlength{\tabcolsep}{10pt}
\caption{Comparison of AttentionMixer and its variants  (mean$\pm$std).}\label{tab:ablation}
\begin{tabular}{llllllll}
    \toprule
    Dataset    & \multicolumn{3}{c}{Beijing Station}    && \multicolumn{3}{c}{Baicheng Station} \\ \cmidrule{2-4} \cmidrule{6-8} 
    Metrics & \mc{RMSE $\downarrow$} & \mc{MAE $\downarrow$} & \mc{R$^2$ $\uparrow$} && \mc{RMSE $\downarrow$} & \mc{MAE $\downarrow$} & \mc{R$^2$ $\uparrow$}\\ \midrule
    w/o SAMP	& 0.0572$\pm$0.0025\sig & 0.0315$\pm$0.0034\sig & 0.8066$\pm$0.0167\sig & &    0.0339$\pm$0.0008\sig & 0.0155$\pm$0.0007\sig & 0.8413$\pm$0.0074\sig\\
    w/o TAMP	& 0.0234$\pm$0.0009\sig & 0.0124$\pm$0.0008\sig & 0.9676$\pm$0.0025\sig  & & 0.0145$\pm$0.0009\sig & 0.0085$\pm$0.0004\sig & 0.9710$\pm$0.0036\sig\\
    w/o SMPR	& 0.0240$\pm$0.0012\sig & 0.0118$\pm$0.0003\sig & 0.9660$\pm$0.0036\sig & & 0.0141$\pm$0.0019 & 0.0073$\pm$0.0008 & 0.9723$\pm$0.0070\sig    \\
    w/o asymmetry & 0.0209$\pm$0.0021 & 0.0104$\pm$0.0008 & 0.9739$\pm$0.0052 & &	0.0137$\pm$0.0013 & 0.0069$\pm$0.0004 & 0.9740$\pm$0.0052\sig  \\
    AttentionMixer & \textbf{0.0203$\pm$0.0008} & \textbf{0.0104$\pm$0.0005} & \textbf{0.9756$\pm$0.0019} & & \textbf{0.0125$\pm$0.0008} & \textbf{0.0066$\pm$0.0003} & \textbf{0.9783$\pm$0.0027}\\
    \bottomrule
  \end{tabular}
\begin{tablenotes}
\item $^*$ marks the variants significantly inferior to AttentionMixer at p-value$<0.05$ over paired sample t-test.   Bolded fonts indicate the best results.
\end{tablenotes}
\end{table*}
\begin{figure*}[]
    \centering
    \subfigure[Performance with different strengths of SMPR $\lambda$ ($\times 10^{-2}$).]{\includegraphics[width=\columnwidth]{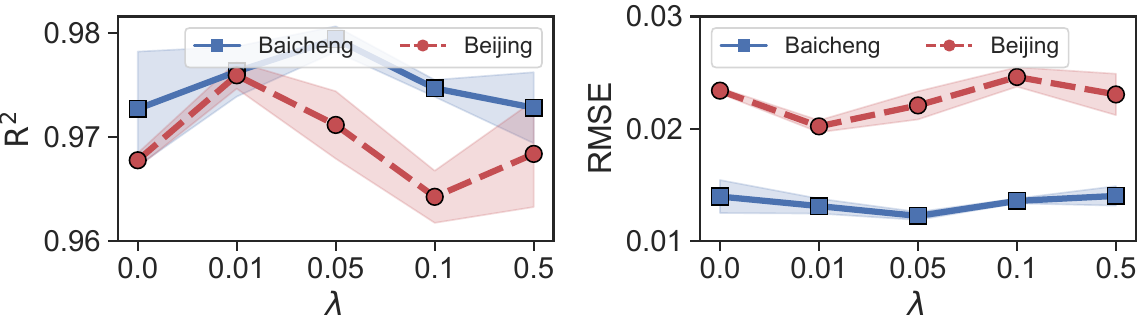}}
    \subfigure[Performance with different rounds of message passing K.]{\includegraphics[width=\columnwidth]{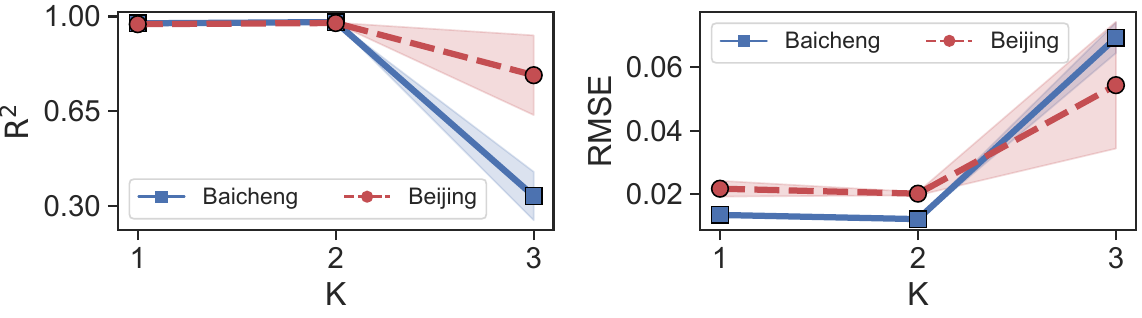}}
    \subfigure[Performance with different settings of window length L.]{\includegraphics[width=\columnwidth]{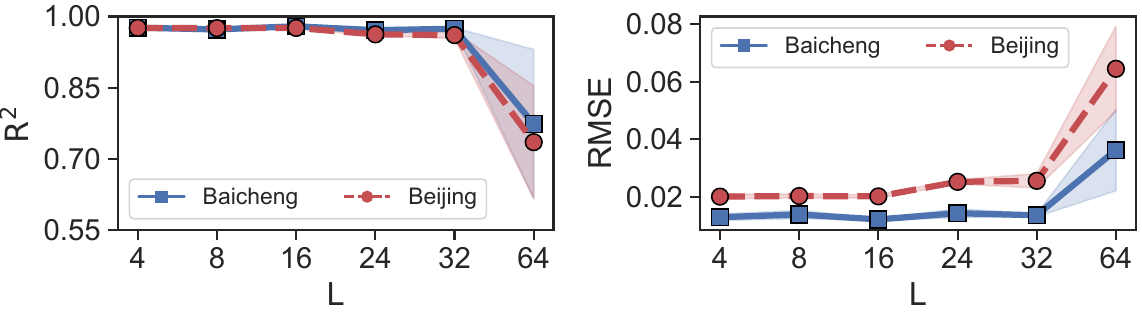}}
    \subfigure[Performance with different settings batch size.]{\includegraphics[width=\columnwidth]{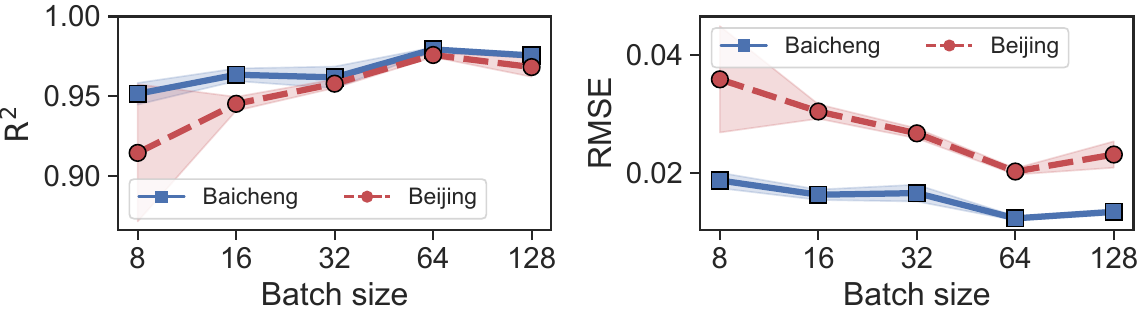}}
    \caption{Parameter sensitivity studies of AttentionMixer, where the marker and dark area indicates the mean value and 90\% confidence interval, respectively.}
    \label{fig:sensi}
\end{figure*}

\begin{figure*}
\centering
\subfigure[$\lambda=0$]{
\includegraphics[scale=0.075, trim=30 0 180 5,clip]{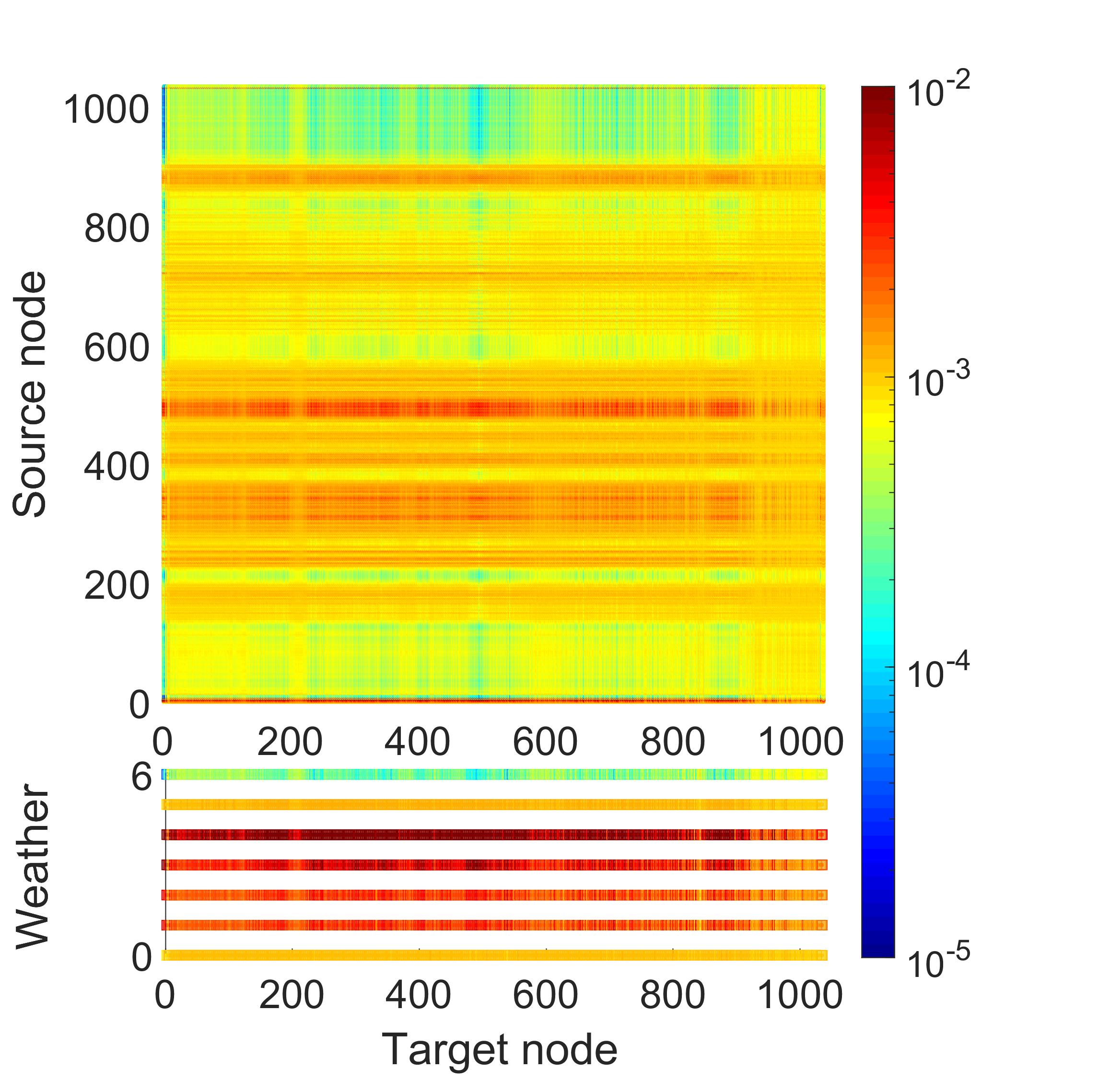}}
\quad
\subfigure[$\lambda=0.03$]{
\includegraphics[scale=0.075, trim=30 0 180 5,clip]{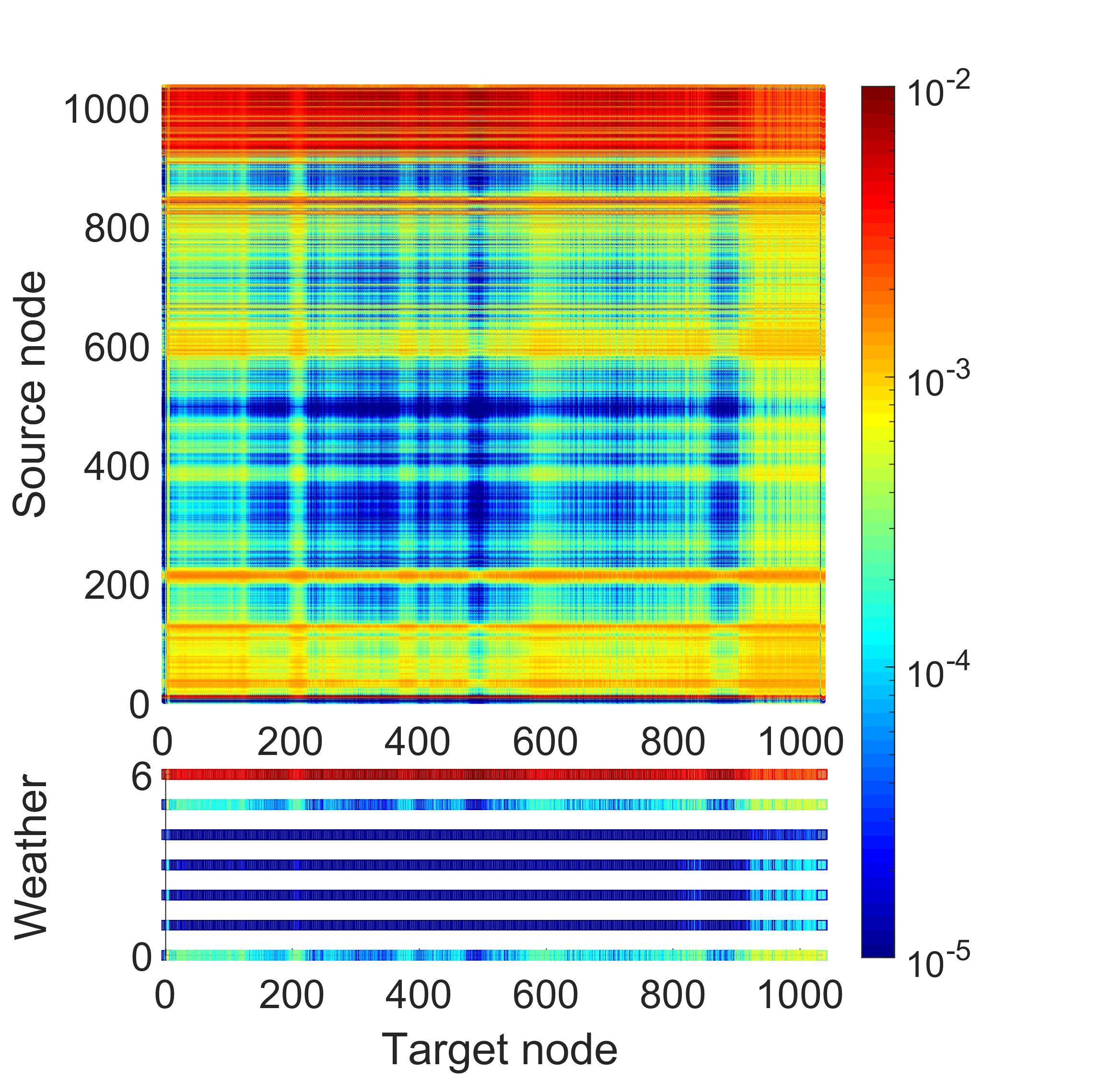}}
\quad
\subfigure[$\lambda=0.1$]{
\includegraphics[scale=0.075, trim=30 0 180 5,clip]{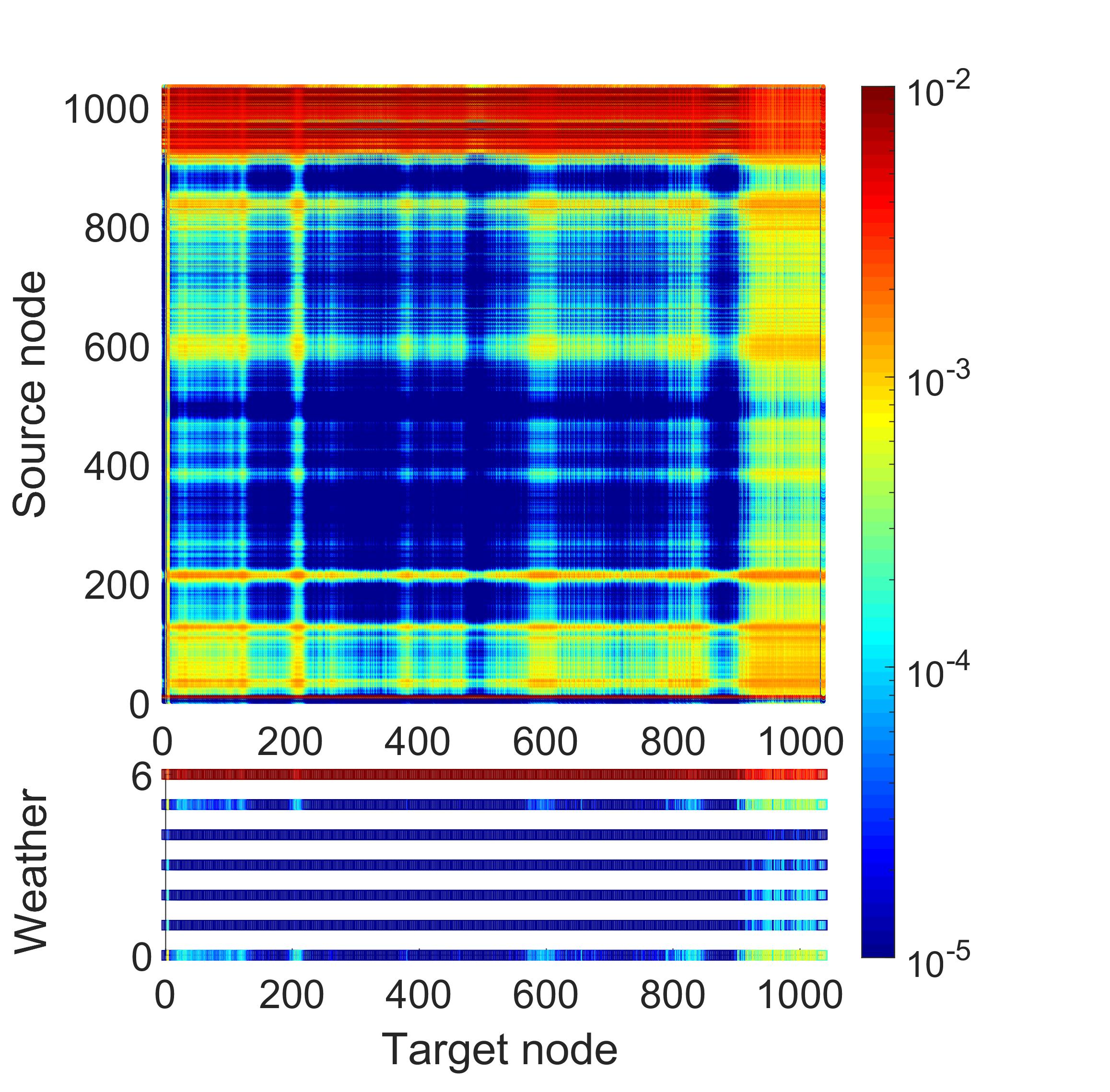}}
\caption{The attention matrix in the first SAMP module with different strengths of SMPR $\lambda$. The weather conditions from bottom to top are temperature, humidity, atmospheric pressure, wind direction, wind speed, precipitation indicator, and amount of precipitation.}\label{explainChart}
\end{figure*}

\begin{table}
\caption{Complexity Analysis of four blocks}
\label{tab:complex}
\resizebox{\columnwidth}{!}{
\setlength{\tabcolsep}{9pt}
\begin{tabular}{llll}
    \toprule
    Blocks               &Complexity                &Sequential Ops                    &Path Length\\
    \midrule
    RNN & $\mathcal{O}(\T\cdot\D^2)$ & $\mathcal{O}(\T)$ & $\mathcal{O}(\T)$ \\
    TCN & $\mathcal{O}(\tau\cdot\T\cdot\D^2)$ & $\mathcal{O}(1)$ & $\mathcal{O}(\log_\tau{\T})$\\
    TAMP& $\mathcal{O}(\T^2\cdot\D)$ & $\mathcal{O}(1)$ & $\mathcal{O}(1)$\\
    SAMP& $\mathcal{O}(\T\cdot\D^2)$ & $\mathcal{O}(1)$ & $\mathcal{O}(1)$\\
    \bottomrule
  \end{tabular}
}
\end{table}

We deconstruct various aspects of AttentionMixer to assess the role of each component in improving accuracy. There are  four variant models devised:
\begin{enumerate}
    \item w/o SAMP: AttentionMixer without the SAMP module;
    \item w/o TAMP: AttentionMixer without the TAMP module;
    \item w/o SMPR: AttentionMixer without SMPR, \ie, $\lambda=0$;
    \item w/o asymmetry: AttentionMixer with undirected message passing graph, \ie, $\mathbf{Q}_{\mathrm{S}}^{k}=\mathbf{K}_{\mathrm{S}}^{k}$ and $\mathbf{Q}_{\mathrm{T}}^{k}=\mathbf{K}_{\mathrm{T}}^{k}$.
\end{enumerate}

The results in \autoref{tab:ablation} reveal the observations as follows.
First, both SAMP and TAMP modules are crucial for achieving overall performance, emphasizing the significance of variate-wise and step-wise correlations in accurate ECP monitoring.
Second, SMPR, which eliminates noisy correlations and reduces overfitting risks, significantly improves AttentionMixer's performance. For instance, it enhances $\mathrm{R}^2$ from 0.966 to 0.975 on the Beijing dataset and reduces RMSE from 0.014 to 0.012 on the Baicheng dataset.
Finally, the inclusion of asymmetry in attention matrices also contributes to improved performance. Notably, on a temporal scale, the latter step is a consequence of the former step, but not vice versa; on a spatial scale, meteorological conditions affect dose-rate measures, but not vice versa.
Hence, such asymmetry exists in nuclear radiation monitoring logs, and disregarding this property diminishes model accuracy.

\subsubsection{Interpretability}
\begin{figure*}
\centering
\subfigure[Inference time on the Intel(R) Xeon(R) Gold 6140 CPU.]{
\includegraphics[width=\columnwidth]{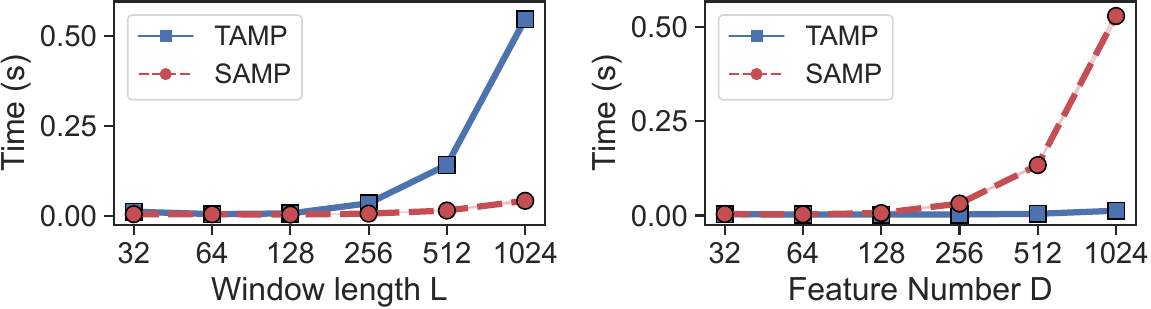}}
\subfigure[Inference time on the Nvidia RTX Titan GPU.]{
\includegraphics[width=0.99\columnwidth]{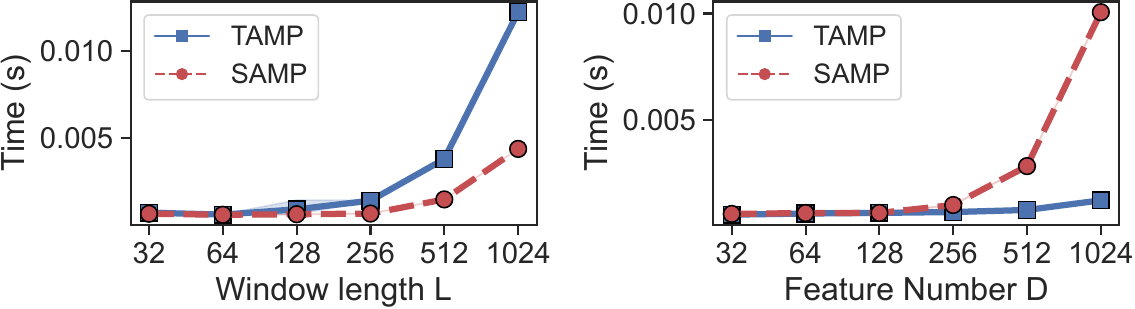}}
\caption{Inference time of the SAMP and TAMP blocks on CPUs (a) and GPUs (b). The default settings of L and D are 16 and 32, respectively. The batch size is set to 64 and the results are mean values of 10 trials.}\label{fig:speed}
\end{figure*}

Interpretability is crucial for trusting AI-based ECP monitors. However, existing solutions, ranging from conventional deep learning models like RNNs and TCNs to standard self-attention methods, suffer from a lack of interpretability. This is predominantly due to their reliance on opaque non-linear transformations or overemphasis on step-wise correlations.
To overcome these challenges, we first propose the SAMP block which models variate-wise correlations to enhance transparency in decision-making, as visualized in \autoref{explainChart}.  
Unlike conventional step-wise attention mechanisms, such variate-wise correlations can be understood and interpreted by experts in ECP monitoring.

Furthermore, a major shortcoming of current attention mechanisms is the density of the attention matrices, as shown in \autoref{explainChart} (a), whereas the physically meaningful correlations  are often sparse. The spurious correlations overshadow the physically meaningful ones, thus hindering interpretability.
SMPR addresses this by sparsifying the attention matrix, which enables significant relationships to stand out, thereby enhancing interpretability as shown in \autoref{explainChart} (b-c).

Moreover, the correctness of model's decision-making process can be validated with expert knowledge. 
In our case study, domain experts are principally concerned with whether the model's understanding of correlations between meteorological conditions and radiation spectrum measurements coheres with existing physical principles.
In particular, precipitation is widely acknowledged to have a pronounced impact on spectrum measurements, since precipitation significantly influences the transport and concentration of radioactive particles, as well as the volume of radon washed from the air to the ground.
Corroborating this, the lower half of \autoref{explainChart} reveals that strong correlations between precipitation and spectrum measurements are maintained, even when the strength of SMPR is increased. 
Similarly, temperature is another influential meteorological variable, aligning with expert understanding: heating causes radon to disperse upwards and away from the sensor network's ground level. Meanwhile, wind speed and direction have marginal effects, a result consistent with the limited fluctuations of these variables in the observed time period and the distinct locations of monitoring stations.
Such alignment between the model's understanding of the data and the experts' knowledge amplifies the model's trustworthiness, thereby augmenting its applicability in real-world ECP monitoring tasks.

\subsection{Parameter Sensitivity Study}\label{sec:param}
We examine hyperparameters that can largely affect model performance: the strength of SMPR ($\lambda$), the round of message passing (K), the window length (L), and the batch size.

SMPR positively contributes to the overall monitoring performance within a wide range of $\lambda$ values. Specifically, in \autoref{fig:sensi}, the value of $\mathrm{R}^2$ increases from 0.968 at $\lambda=0$ to 0.976 at $\lambda=0.0001$. However, assigning a large weight to SMPR can have a detrimental effect on the performance, with $\mathrm{R}^2$ dropping to 0.969 at $\lambda=0.005$.
It is speculated that over-emphasizing SMPR in a multitask learning framework could make it challenging to minimize forecast errors effectively.

A single round of message passing appears adequate to obtain a comprehensive representation. Additional rounds of message passing, according to~\autoref{fig:sensi}, provide limited benefits and may even lead to performance drop due to overfitting and optimization difficulties.
The performance is insensitive to window length (L) within a larger range, indicating that modeling overly long-term step-wise relationship is unnecessary for ECP monitoring, whereas variate-wise relationships seem crucial for achieving accurate monitoring. Concurrently, larger batch sizes can stabilize the training process and improve the final results within a wide range.
\subsection{Complexity Analysis}\label{sec:complex}

In this section, we compare the computational costs among different sequential models following~\cite{Vaswani2017}. The methods are employed to transform one sequence $(\mathbf{x}_1,...,\mathbf{x}_\T)$ to another sequence $(\mathbf{z}_1,...,\mathbf{z}_\T)$ with equal length $\T$ and feature number $\D$. 
Three metrics are considered, including complexity, sequential ops, and path length, which respectively measure the amount of floating-point operations, non-parallelizable operations, and the minimum number of layers to model the relationships between any time steps. 

According to \autoref{tab:complex}, RNNs exhibit inefficiencies in training and inference due to sequential operations and path length. TCNs, with a kernel length $\tau$, enable full parallelization; however, the reliance on multiple stacked TCN layers to model global relationships hinders efficiency for long-term dependencies. TAMPs, i.e., the vanilla self-attention block, demonstrate advantages in terms of parallelization and path length, but the complexity of $\mathcal{O}(\T^2)$ poses challenges in handling large window length~\cite{bigbird}. The proposed SAMP module enjoys the superiority of parallelization and path length while reducing the complexity to $\mathcal{O}(\T)$ in relation to window length.

\autoref{fig:speed} reports the inference time of the SAMP and TAMP modules for 64 samples, where the inference cost satisfies the requirements of ECP monitoring. The observed evolution of inference costs aligns with the analytical results presented in \autoref{tab:complex}.  
One potential drawback of the SAMP module is its second-order complexity in relation to $\mathcal{\D}$. However, this issue can be mitigated by mapping the input data into a lower-dimensional space before the SAMP block.

\section{Conclusion}\label{sec:conclusion}
This study proposes a novel AttentionMixer framework to address the current limitations of ECP monitoring methods in terms of accuracy and interpretability, with the goal of providing a trustworthy process monitoring framework. The framework employs spatial and temporal adaptive message passing blocks to build a comprehensive representation of ECP logs.
A sparse message passing regularizer is utilized to eliminate spurious and noisy correlations.
By explicitly modeling variate-wise correlations and effectively eliminating spurious correlations, AttentionMixer improves accuracy, interpretability, and thereby trustworthiness of ECP monitoring.

There are some avenues remained to be investigate. First, it is promising to explore the role and flexibility of techniques in vanilla Transformers, such as patching~\cite{nie2022time}, sparse attentions~\cite{bigbird}, and adversarial training~\cite{wu2020adversarial}, within the AttentionMixer framework, particularly in the SAMP block that is constructed based on variate-wise correlations. This exploration is expected to further improve the monitoring quality of AttentionMixer. The second future work is to deploy the proposed model on our online monitoring board\footnotemark[1] to serve nuclear power plants nationwide, which will enable us to evaluate the performance of the framework under a broader range of conditions and further refine its effectiveness.
\bibliographystyle{ieeetr}
\bibliography{Bibliography/main.bib}\ 

\begin{thebibliography}{10}

\bibitem{wu2016embodied}
X.~Wu, X.-H. Xia, G.~Chen, X.~Wu, and B.~Chen, ``Embodied energy analysis for
  coal-based power generation system-highlighting the role of indirect energy
  cost,'' {\em Appl. Energy}, vol.~184, pp.~936--950, 2016.

\bibitem{safe1}
F.-C. Chen and M.~R. Jahanshahi, ``Nb-cnn: Deep learning-based crack detection
  using convolutional neural network and naïve bayes data fusion,'' {\em IEEE
  Trans. Ind. Electron.}, vol.~65, no.~5, pp.~4392--4400, 2018.

\bibitem{safe2}
M.~Embrechts and S.~Benedek, ``Hybrid identification of nuclear power plant
  transients with artificial neural networks,'' {\em IEEE Trans. Ind.
  Electron.}, vol.~51, no.~3, pp.~686--693, 2004.

\bibitem{liang2022advances}
W.~Liang, G.~A. Tadesse, D.~Ho, L.~Fei-Fei, M.~Zaharia, C.~Zhang, and J.~Zou,
  ``Advances, challenges and opportunities in creating data for trustworthy
  ai,'' {\em Nat. Mach. Intell.}, vol.~4, no.~8, pp.~669--677, 2022.

\bibitem{trustworthytai1}
A.~Rawal, J.~McCoy, D.~B. Rawat, B.~M. Sadler, and R.~S. Amant, ``Recent
  advances in trustworthy explainable artificial intelligence: Status,
  challenges, and perspectives,'' {\em {IEEE} Trans. Artif. Intell.}, vol.~3,
  no.~6, pp.~852--866, 2022.

\bibitem{trustworthytai2}
K.~A. Crockett, E.~Colyer, L.~Gerber, and A.~Latham, ``Building trustworthy
  {AI} solutions: {A} case for practical solutions for small businesses,'' {\em
  {IEEE} Trans. Artif. Intell.}, vol.~4, no.~4, pp.~778--791, 2023.

\bibitem{wang2023out}
H.~Wang, K.~Kuang, L.~Lan, Z.~Wang, W.~Huang, F.~Wu, and W.~Yang,
  ``Out-of-distribution generalization with causal feature separation,'' {\em
  IEEE Trans. Knowl. Data Eng.}, 2023.

\bibitem{trustworthytai4}
Z.~Chen, F.~Silvestri, G.~Tolomei, J.~Wang, H.~Zhu, and H.~Ahn, ``Explain the
  explainer: Interpreting model-agnostic counterfactual explanations of a deep
  reinforcement learning agent,'' {\em {IEEE} Trans. Artif. Intell.}, 2022.

\bibitem{trustworthytai3}
A.~Takiddin, M.~Ismail, R.~Atat, K.~R. Davis, and E.~Serpedin, ``Robust graph
  autoencoder-based detection of false data injection attacks against data
  poisoning in smart grids,'' {\em {IEEE} Trans. Artif. Intell.}, 2023.

\bibitem{zhang2023robust}
Z.~Zhang, Q.~Dai, X.~Chen, Z.~Dong, and R.~Tang, ``Robust causal inference for
  recommender system to overcome noisy confounders,'' in {\em {SIGIR}},
  pp.~2349--2353, {ACM}, 2023.

\bibitem{wang2022entire}
H.~Wang, Z.~Chen, J.~Fan, Y.~Huang, W.~Liu, and X.~Liu, ``Entire space
  counterfactual learning: Tuning, analytical properties and industrial
  applications,'' {\em arXiv preprint arXiv:2210.11039}, 2022.

\bibitem{nightingale2022ai}
S.~J. Nightingale and H.~Farid, ``Ai-synthesized faces are indistinguishable
  from real faces and more trustworthy,'' {\em P. NATL. A. SCI.}, vol.~119,
  no.~8, p.~e2120481119, 2022.

\bibitem{sheth2021knowledge}
A.~Sheth, M.~Gaur, K.~Roy, and K.~Faldu, ``Knowledge-intensive language
  understanding for explainable ai,'' {\em IEEE Internet Computing}, vol.~25,
  no.~5, pp.~19--24, 2021.

\bibitem{DBLP:conf/icml/LiZCGLW23}
H.~Li, C.~Zheng, Y.~Cao, Z.~Geng, Y.~Liu, and P.~Wu, ``Trustworthy policy
  learning under the counterfactual no-harm criterion,'' in {\em {ICML}},
  vol.~202 of {\em Proceedings of Machine Learning Research}, pp.~20575--20598,
  {PMLR}, 2023.

\bibitem{DBLP:conf/kdd/LiZWKL023}
H.~Li, C.~Zheng, P.~Wu, K.~Kuang, Y.~Liu, and P.~Cui, ``Who should be given
  incentives? counterfactual optimal treatment regimes learning for
  recommendation,'' in {\em {KDD}}, pp.~1235--1247, {ACM}, 2023.

\bibitem{wuite}
A.~Wu, J.~Yuan, K.~Kuang, B.~Li, R.~Wu, Q.~Zhu, Y.~Zhuang, and F.~Wu,
  ``Learning decomposed representations for treatment effect estimation,'' {\em
  {IEEE} Trans. Knowl. Data Eng.}, vol.~35, no.~5, pp.~4989--5001, 2023.

\bibitem{wustable}
A.~Wu, K.~Kuang, R.~Xiong, B.~Li, and F.~Wu, ``Stable estimation of
  heterogeneous treatment effects,'' in {\em {ICML}}, vol.~202 of {\em
  Proceedings of Machine Learning Research}, pp.~37496--37510, {PMLR}, 2023.

\bibitem{DBLP:conf/kdd/YangCL0023}
M.~Yang, X.~Cai, F.~Liu, W.~Zhang, and J.~Wang, ``Specify robust causal
  representation from mixed observations,'' in {\em {KDD}}, pp.~2978--2987,
  {ACM}, 2023.

\bibitem{DBLP:conf/cvpr/YangLCSHW21}
M.~Yang, F.~Liu, Z.~Chen, X.~Shen, J.~Hao, and J.~Wang, ``Causalvae:
  Disentangled representation learning via neural structural causal models,''
  in {\em {CVPR}}, pp.~9593--9602, Computer Vision Foundation / {IEEE}, 2021.

\bibitem{DBLP:conf/kdd/0001KCYGH023}
H.~Wang, K.~Kuang, H.~Chi, L.~Yang, M.~Geng, W.~Huang, and W.~Yang, ``Treatment
  effect estimation with adjustment feature selection,'' in {\em {KDD}},
  pp.~2290--2301, {ACM}, 2023.

\bibitem{DBLP:conf/kdd/00010YWXR0K22}
H.~Wang, W.~Yang, L.~Yang, A.~Wu, L.~Xu, J.~Ren, F.~Wu, and K.~Kuang,
  ``Estimating individualized causal effect with confounded instruments,'' in
  {\em {KDD}}, pp.~1857--1867, {ACM}, 2022.

\bibitem{DBLP:conf/sigir/WangZL0F023}
W.~Wang, Y.~Zhang, H.~Li, P.~Wu, F.~Feng, and X.~He, ``Causal recommendation:
  Progresses and future directions,'' in {\em {SIGIR}}, pp.~3432--3435, {ACM},
  2023.

\bibitem{wutest}
A.~Wu, H.~Li, K.~Kuang, K.~Zhang, and F.~Wu, ``Hierarchical topological
  ordering with conditional independence test for limited time series,'' {\em
  CoRR}, vol.~abs/2308.08148, 2023.

\bibitem{DBLP:conf/iclr/LiLZ023}
H.~Li, Y.~Lyu, C.~Zheng, and P.~Wu, ``{TDR-CL:} targeted doubly robust
  collaborative learning for debiased recommendations,'' in {\em {ICLR}},
  OpenReview.net, 2023.

\bibitem{li2022multiple}
H.~Li, Q.~Dai, Y.~Li, Y.~Lyu, Z.~Dong, P.~Wu, and X.-H. Zhou, ``Multiple robust
  learning for recommendation,'' {\em arXiv preprint arXiv:2207.10796}, 2022.

\bibitem{li2023stabledr}
H.~Li, C.~Zheng, and P.~Wu, ``Stabledr: Stabilized doubly robust learning for
  recommendation on data missing not at random,'' in {\em ICLR}, 2023.

\bibitem{wangescm}
H.~Wang, T.~Chang, T.~Liu, J.~Huang, Z.~Chen, C.~Yu, R.~Li, and W.~Chu,
  ``{ESCM2:} entire space counterfactual multi-task model for post-click
  conversion rate estimation,'' in {\em {SIGIR}}, pp.~363--372, {ACM}, 2022.

\bibitem{DBLP:conf/sigir/ZhangDCDT23}
Z.~Zhang, Q.~Dai, X.~Chen, Z.~Dong, and R.~Tang, ``Robust causal inference for
  recommender system to overcome noisy confounders,'' in {\em {SIGIR}},
  pp.~2349--2353, {ACM}, 2023.

\bibitem{DBLP:conf/icml/LiXZ0023}
H.~Li, Y.~Xiao, C.~Zheng, P.~Wu, and P.~Cui, ``Propensity matters: Measuring
  and enhancing balancing for recommendation,'' in {\em {ICML}}, vol.~202 of
  {\em Proceedings of Machine Learning Research}, pp.~20182--20194, {PMLR},
  2023.

\bibitem{DBLP:conf/sigir/YangWT23}
M.~Yang, J.~Wang, and J.~Ton, ``Rectifying unfairness in recommendation
  feedback loop,'' in {\em {SIGIR}}, pp.~28--37, {ACM}, 2023.

\bibitem{markus2021role}
A.~F. Markus, J.~A. Kors, and P.~R. Rijnbeek, ``The role of explainability in
  creating trustworthy artificial intelligence for health care: a comprehensive
  survey of the terminology, design choices, and evaluation strategies,'' {\em
  J. Biomed. Inform.}, vol.~113, p.~103655, 2021.

\bibitem{floridi2019establishing}
L.~Floridi, ``Establishing the rules for building trustworthy ai,'' {\em Nat.
  Mach. Intell.}, vol.~1, no.~6, pp.~261--262, 2019.

\bibitem{yuan1}
X.~Yuan, L.~Li, and Y.~Wang, ``Nonlinear dynamic soft sensor modeling with
  supervised long short-term memory network,'' {\em IEEE Trans. Ind.
  Informat.}, vol.~16, no.~5, pp.~3168--3176, 2019.

\bibitem{yuan2}
X.~Yuan, J.~Zhou, B.~Huang, Y.~Wang, C.~Yang, and W.~Gui, ``Hierarchical
  quality-relevant feature representation for soft sensor modeling: A novel
  deep learning strategy,'' {\em IEEE Trans. Ind. Informat.}, vol.~16, no.~6,
  pp.~3721--3730, 2019.

\bibitem{xiao2022distributed}
C.~Xiao, W.~Han, W.~Shao, and D.~Zhao, ``Distributed semisupervised hmm for
  dynamic inferential sensor development,'' {\em IEEE Sens. J}, vol.~23, no.~3,
  pp.~2737--2749, 2022.

\bibitem{li2022novel}
H.~Li, W.~Wang, Z.~Liu, Y.~Niu, H.~Wang, S.~Zhao, Y.~Liao, W.~Yang, and X.~Liu,
  ``A novel locality-sensitive hashing relational graph matching network for
  semantic textual similarity measurement,'' {\em Expert Syst. Appl.},
  vol.~207, p.~117832, 2022.

\bibitem{fanlearnable}
J.~Fan, Y.~Zhuang, Y.~Liu, H.~Jianye, B.~Wang, J.~Zhu, H.~Wang, and S.-T. Xia,
  ``Learnable behavior control: Breaking atari human world records via
  sample-efficient behavior selection,'' in {\em ICLR}, 2023.

\bibitem{liu2023novel}
Z.~Liu, H.~Li, H.~Wang, Y.~Liao, X.~Liu, and G.~Wu, ``A novel pipelined
  end-to-end relation extraction framework with entity mentions and contextual
  semantic representation,'' {\em Expert Syst. Appl.}, p.~120435, 2023.

\bibitem{Chandrakar2017}
A.~Chandrakar, D.~Datta, A.~K. Nayak, and G.~Vinod, ``{Statistical analysis of
  a time series relevant to passive systems of nuclear power plants},'' {\em
  Int. J. Syst. Assur. Eng. Manag.}, vol.~8, no.~1, pp.~89--108, 2017.

\bibitem{xgb}
J.~I. Aizpurua, S.~D.~J. McArthur, B.~G. Stewart, B.~Lambert, J.~G. Cross, and
  V.~M. Catterson, ``Adaptive power transformer lifetime predictions through
  machine learning and uncertainty modeling in nuclear power plants,'' {\em
  IEEE Trans. Ind. Electron.}, vol.~66, no.~6, pp.~4726--4737, 2019.

\bibitem{li2023virtual}
Y.~Li, W.~Han, W.~Shao, and D.~Zhao, ``Virtual sensing for dynamic industrial
  process based on localized linear dynamical system models with time-delay
  optimization,'' {\em ISA Trans.}, vol.~133, pp.~505--517, 2023.

\bibitem{chenmode}
Z.~Chen, L.~Ding, Z.~Chu, Y.~Qi, J.~Huang, and H.~Wang, ``Monotonic neural
  ordinary differential equation: Time-series forecasting for cumulative
  data,'' in {\em CIKM}, 2023.

\bibitem{chensvgd}
Z.~Chen, L.~Ding, J.~Huang, Z.~Chu, Q.~Dai, and H.~Wang, ``Unsupervised anomaly
  detection \& diagnosis: A stein variational gradient descent approach,'' in
  {\em CIKM}, 2023.

\bibitem{dai2022incremental}
Q.~Dai, C.~Zhao, and B.~Huang, ``Incremental variational bayesian gaussian
  mixture model with decremental optimization for distribution accommodation
  and fine-scale adaptive process monitoring,'' {\em IEEE Trans. Cybern.},
  2022.

\bibitem{dai2023variational}
Q.~Dai, C.~Zhao, and S.~Zhao, ``Variational bayesian student'st mixture model
  with closed-form missing value imputation for robust process monitoring of
  low-quality data,'' {\em IEEE Trans. Cybern.}, 2023.

\bibitem{jiang2022data}
Q.~Jiang, Z.~Wang, S.~Yan, and Z.~Cao, ``Data-driven soft sensing for batch
  processes using neural network-based deep quality-relevant representation
  learning,'' {\em {IEEE} Trans. Artif. Intell.}, 2022.

\bibitem{huangbmoe}
Y.~Huang, H.~Wang, Z.~Liu, L.~Pan, H.~Li, and X.~Liu, ``Modeling task
  relationships in multi-variate soft sensor with balanced
  mixture-of-experts,'' {\em IEEE Trans. Ind. Informat.}, pp.~1--9, 2022.

\bibitem{wang2020remaining}
H.~Wang, M.~Peng, R.~Xu, A.~Ayodeji, and H.~Xia, ``Remaining useful life
  prediction based on improved temporal convolutional network for nuclear power
  plant valves,'' {\em Front. Energy Res.}, vol.~8, p.~296, 2020.

\bibitem{Wu2020}
Z.~Wu, S.~Pan, G.~Long, J.~Jiang, X.~Chang, and C.~Zhang, ``{Connecting the
  Dots: Multivariate Time Series Forecasting with Graph Neural Networks},'' in
  {\em SIGKDD}, pp.~753--763, 2020.

\bibitem{jumper2021highly}
J.~Jumper, R.~Evans, A.~Pritzel, T.~Green, M.~Figurnov, O.~Ronneberger,
  K.~Tunyasuvunakool, R.~Bates, {\em et~al.}, ``Highly accurate protein
  structure prediction with alphafold,'' {\em Nature}, vol.~596, no.~7873,
  pp.~583--589, 2021.

\bibitem{chen2021knowledge}
Z.~Chen and Z.~Ge, ``Knowledge automation through graph mining, convolution and
  explanation framework: a soft sensor practice,'' {\em IEEE Trans. Ind.
  Informat.}, 2021.

\bibitem{ma2021deep}
Y.~Ma and J.~Tang, {\em Deep Learning on Graphs}.
\newblock Cambridge University Press, 2021.

\bibitem{Vaswani2017}
A.~Vaswani, N.~Shazeer, N.~Parmar, J.~Uszkoreit, L.~Jones, A.~N. Gomez,
  {\L}.~Kaiser, and I.~Polosukhin, ``Attention is all you need,'' {\em
  NeurIPS}, vol.~30, 2017.

\bibitem{zhou2022fedformer}
T.~Zhou, Z.~Ma, Q.~Wen, X.~Wang, L.~Sun, and R.~Jin, ``Fedformer: Frequency
  enhanced decomposed transformer for long-term series forecasting,'' in {\em
  ICML}, pp.~27268--27286, PMLR, 2022.

\bibitem{zhou2021informer}
H.~Zhou, S.~Zhang, J.~Peng, S.~Zhang, J.~Li, H.~Xiong, and W.~Zhang,
  ``Informer: Beyond efficient transformer for long sequence time-series
  forecasting,'' in {\em AAAI}, vol.~35, pp.~11106--11115, 2021.

\bibitem{liu2021pyraformer}
S.~Liu, H.~Yu, C.~Liao, J.~Li, W.~Lin, A.~X. Liu, and S.~Dustdar, ``Pyraformer:
  Low-complexity pyramidal attention for long-range time series modeling and
  forecasting,'' in {\em ICLR}, 2021.

\bibitem{li2019enhancing}
S.~Li, X.~Jin, Y.~Xuan, X.~Zhou, W.~Chen, Y.-X. Wang, and X.~Yan, ``Enhancing
  the locality and breaking the memory bottleneck of transformer on time series
  forecasting,'' in {\em NeurIPS}, vol.~32, 2019.

\bibitem{wu2020adversarial}
S.~Wu, X.~Xiao, Q.~Ding, P.~Zhao, Y.~Wei, and J.~Huang, ``Adversarial sparse
  transformer for time series forecasting,'' {\em NeurIPS}, vol.~33,
  pp.~17105--17115, 2020.

\bibitem{nie2022time}
Y.~Nie, N.~H. Nguyen, P.~Sinthong, and J.~Kalagnanam, ``A time series is worth
  64 words: Long-term forecasting with transformers,'' in {\em ICLR}, 2023.

\bibitem{bigbird}
M.~Zaheer, G.~Guruganesh, K.~A. Dubey, J.~Ainslie, C.~Alberti,
  S.~Onta{\~{n}}{\'{o}}n, P.~Pham, A.~Ravula, Q.~Wang, L.~Yang, and A.~Ahmed,
  ``Big bird: Transformers for longer sequences,'' in {\em NeurIPS}, vol.~33,
  pp.~17283--17297, 2020.

\bibitem{use1}
P.~Ball {\em et~al.}, ``The start-ups chasing clean, carbon-free fusion
  energy,'' {\em Nature}, vol.~599, no.~7885, pp.~362--366, 2021.

\bibitem{use2}
R.~Y. Cui, N.~Hultman, D.~Cui, H.~McJeon, S.~Yu, M.~R. Edwards, A.~Sen,
  K.~Song, C.~Bowman, L.~Clarke, {\em et~al.}, ``A plant-by-plant strategy for
  high-ambition coal power phaseout in china,'' {\em Nat. Commun.}, vol.~12,
  no.~1, pp.~1--10, 2021.

\bibitem{roth2017going}
M.~B. Roth and P.~Jaramillo, ``Going nuclear for climate mitigation: An
  analysis of the cost effectiveness of preserving existing us nuclear power
  plants as a carbon avoidance strategy,'' {\em Energy}, vol.~131, pp.~67--77,
  2017.

\bibitem{weather1}
J.~Hirouchi, S.~Hirao, J.~Moriizumi, H.~Yamazawa, and A.~Suzuki, ``Estimation
  of infiltration and surface run-off characteristics of radionuclides from
  gamma dose rate change after rain,'' {\em J. Nucl. Sci. Technol.}, vol.~51,
  no.~1, pp.~48--55, 2014.

\bibitem{weather3}
J.~F. Mercier, B.~L. Tracy, R.~D'Amours, F.~Chagnon, I.~Hoffman, E.~P. Korpach,
  S.~Johnson, and R.~K. Ungar, ``{Increased environmental gamma-ray dose rate
  during precipitation: a strong correlation with contributing air mass},''
  {\em J. Environ. Radioact.}, vol.~100, no.~7, pp.~527--533, 2009.

\bibitem{Chan2017}
Y.~K. Chan and Y.~C. Tsai, ``{Multiple regression approach to predict
  turbine-generator output for Chinshan nuclear power plant},'' {\em
  Kerntechnik}, vol.~82, no.~1, pp.~24--30, 2017.

\bibitem{Tang2012}
L.~Tang, L.~Yu, S.~Wang, J.~Li, and S.~Wang, ``{A novel hybrid ensemble
  learning paradigm for nuclear energy consumption forecasting},'' {\em Appl.
  Energy}, vol.~93, pp.~432--443, 2012.

\bibitem{rf}
S.~Grape, E.~Branger, Z.~Elter, and L.~P. Balkest{\aa}hl, ``Determination of
  spent nuclear fuel parameters using modelled signatures from non-destructive
  assay and random forest regression,'' {\em Nucl. Instrum. Methods Phys. Res.,
  Sect. A}, vol.~969, p.~163979, 2020.

\bibitem{Choi2020}
J.~Choi and S.~J. Lee, ``{Consistency Index-Based Sensor Fault Detection System
  for Nuclear Power Plant Emergency Situations Using an LSTM Network},'' {\em
  Sensors}, vol.~20, no.~6, p.~1651, 2020.

\bibitem{Zhang2018}
J.~Zhang, Z.~Pan, W.~Bai, and X.~Zhou, ``Pressurizer water level reconstruction
  for nuclear power plant based on gru,'' in {\em IMCCC}, pp.~1675--1679, 2018.

\end{thebibliography}

\end{document}